\documentclass[journal]{IEEEtran}

\ifCLASSINFOpdf
  \usepackage{microtype}
  \usepackage{graphicx}
  \usepackage{subfigure}
  \usepackage{booktabs} % for professional tables
  \usepackage{amsmath}
  \usepackage{amssymb}
  \usepackage{bm}
  \usepackage{multirow}
  \usepackage{verbatim}
  \usepackage{makecell}
  \usepackage{authblk}
  \usepackage{bbding}
  \usepackage{color}
  \usepackage[colorlinks,linkcolor=red]{hyperref}
\else
\fi

\begin{document}
\title{SPU-Net: Self-Supervised Point Cloud Upsampling by Coarse-to-Fine Reconstruction with Self-Projection Optimization}

\author{Xinhai Liu,
        Xinchen Liu,
        Yu-Shen Liu,
        Zhizhong Han
\IEEEcompsocitemizethanks{
\IEEEcompsocthanksitem Xinhai Liu and Xinchen Liu are with the School of Software, BNRist, Tsinghua University, Beijing, China. E-mail: lxh17@mails.tsinghua.edu.cn; lxc19@mails.tsinghua.edu.cn.
\IEEEcompsocthanksitem Yu-Shen Liu is with the School of Software, BNRist, Tsinghua University, Beijing, China. E-mail: liuyushen@tsinghua.edu.cn. Yu-Shen Liu is the corresponding author.
\IEEEcompsocthanksitem Zhizhong Han is with the Department of Computer Science, Wayne State University, USA. E-mail: h312h@wayne.edu.}
\thanks{This work was supported by National Key R\&D Program of China (2020YFF0304100), the National Natural Science Foundation of China (62072268), and in part by Tsinghua-Kuaishou Institute of Future Media Data. Code is available at \url{https://github.com/liuxinhai/SPU-Net}.}}

% \markboth{Journal of \LaTeX\ Class Files,~Vol.~14, No.~8, August~2015}%
% {Shell \MakeLowercase{\textit{et al.}}: Bare Demo of IEEEtran.cls for IEEE Journals}

\maketitle

\begin{abstract}
  The task of point cloud upsampling aims to acquire dense and uniform point sets from sparse and irregular point sets.
  Although significant progress has been made with deep learning models, state-of-the-art methods require ground-truth dense point sets as the supervision, which makes them limited to be trained under synthetic paired training data and not suitable to be under real-scanned sparse data. 
  However, it is expensive and tedious to obtain large numbers of paired sparse-dense point sets as supervision from real-scanned sparse data.
  To address this problem, we propose a self-supervised point cloud upsampling network, named SPU-Net, to capture the inherent upsampling patterns of points lying on the underlying object surface.
  Specifically, we propose a coarse-to-fine reconstruction framework, which contains two main components: point feature extraction and point feature expansion, respectively. 
  In the point feature extraction, we integrate the self-attention module with the graph convolution network (GCN) to capture context information inside and among local regions simultaneously.
  In the point feature expansion, we introduce a hierarchically learnable folding strategy to generate upsampled point sets with learnable 2D grids.
  Moreover, to further optimize the noisy points in the generated point sets, we propose a novel self-projection optimization associated with uniform and reconstruction terms as a joint loss to facilitate the self-supervised point cloud upsampling.
  We conduct various experiments on both synthetic and real-scanned datasets, and the results demonstrate that we achieve comparable performances to state-of-the-art supervised methods.
\end{abstract}


\IEEEpeerreviewmaketitle
\section{Introduction}
Point cloud, as one of the most concise 3D representations, has drawn increasing research attention due to its convenient access from various popular depth sensors, such as LiDARs and RGB-D cameras.
However, raw point clouds obtained from devices are usually sparse, noisy, and non-uniform, which leads to a massive challenge for deep neural networks to deal with such irregular data directly.
Given a sparse, noisy, and non-uniform point cloud, the task of upsampling aims to generate a dense and uniform point set as a trusted representation of the underlying object surface.
Therefore, point cloud upsampling, as an amended operation, is meaningful for many downstream applications like rendering, analysis, reconstruction and other general processing.

Current point cloud upsampling methods with deep learning, such as PU-Net \cite{yu2018pu}, MPU \cite{yifan2019patch}, PU-GAN \cite{li2019pu}, PUGeo-Net \cite{qian2020pugeo} and Dis-PU \cite{li2021point}, have achieved outperforming results on some synthetic datasets such as ShapeNet \cite{chang2015shapenet} and VisionAir repository \cite{Visionair-dataset}.
However, these methods usually require ground-truth dense point sets as the supervision to train the neural network.
And the supervision is usually constructed by sampling on synthetic CAD models from publicly available datasets \cite{li2019pu,yu2018pu}, which is unavailable for the real-scanned data.
Due to the absence of paired ground-truth dense point sets, the aforementioned methods cannot be trained under the real-scanned datasets such as ScanNet \cite{dai2017scannet} and KITTI \cite{geiger2013vision}.
In addition, when data distributions from synthetic object data do not match those from real scans, upsampling networks trained on synthetic object data do not generalize well to real (sparse) scans.
For example, the supervised methods trained on synthetic object data, such as PU-Net \cite{yu2018pu} and PU-GAN \cite{li2019pu}, easily changed the origin topological structure of underlying object surface on the real-scanned data or scene point clouds.
Therefore, it is promising to propose a self-supervised point cloud upsampling method, which does not require dense point sets as the supervision and can keep the original data distributions.

Recently, some unsupervised image super-resolution methods \cite{yuan2018unsupervised,shaham2019singan} have been proposed and achieved outperforming performances in generating high-resolution images.
However, due to the irregular and unordered nature of point clouds, it is non-trivial to directly apply these image super-resolution methods to the unsupervised point cloud upsampling.
Specifically, there are two challenges in the unsupervised point cloud upsampling with deep learning models.
(1) \textit{How to establish practical self-supervised information without the supervision of dense point sets?}
Previous methods, such as PU-GAN \cite{li2019pu} and L2G-AE \cite{liu2019l2g}, first generate a dense point set with deep networks and then downsample the dense point set back into a sparse point set, where some supervision is usually applied to the sparse point sets.
However, there is no direct supervision for the dense point sets in the unsupervised upsampling task, which makes it difficult to capture the inherent upsampling patterns.
To resolve this issue, we propose a coarse-to-fine reconstruction framework to formulate the self-supervised point cloud upsampling. 
Specifically, we first downsample the input patch into some coarse patches and then capture the inherent upsampling patterns by reconstructing the input patch itself from the coarse patch.
Next, a dense patch can be obtained by aggregating multiple fine patches, which follow the distribution of the input patch.
(2) \textit{The point clouds upsampled by deep networks should be a faithful representation of the underlying object surface.}
Due to the irregular nature of the point cloud and network bias in the generation of dense point clouds from sparse ones, it is inevitable to bring some noisy points around the underlying surface when generating the upsampled dense point set, especially for unsupervised upsampling methods.
Therefore, some specific loss functions are needed to constrain the spatial distribution of generated points, including uniformity and flatness.
To resolve these challenges,  we propose a  novel self-projection optimization to constrain the noisy points around the underlying surface to the surface itself, and associate it with uniform and reconstruction terms as a joint loss to facilitate the generation of upsampled points.

To address the above challenges, we propose the coarse-to-fine reconstruction strategy to explore the upsampling patterns with only sparse point sets.
In our approach, we introduce two key components to support the coarse-to-fine point upsampling, named point feature extraction and point feature expansion, respectively. 
In the point feature extraction module, we integrate self-attention with the graph convolutional network (GCN) to fully capture the spatial context of points within local regions.
Furthermore, in the point feature expansion, a hierarchical folding operation with learnable grids is proposed to expand the point features gradually. 
And to refine the final distribution of generated dense points, a joint loss function is designed to constrain multiple geometry attributes, including uniformity, noise, and overall shape.
In general, our contributions are summarised as follows.
\begin{itemize}
    \item We propose a self-supervised point cloud upsampling network (SPU-Net) which can be trained without the supervision of 3D ground-truth dense point clouds. SPU-Net repeatedly upsamples from downsampled patches, which is not restricted by the paired training data and can preserve the original data distribution. 
    \item We propose a coarse-to-fine reconstruction framework to capture the inherent upsampling patterns inside local patches, which introduces a novel self-supervision way to learn to upsample point clouds.
    % contains two main components: point feature extraction and point feature expansion.
    % In the point feature extraction, we integrate self-attention with graph convolution network (GCN) to simultaneously capture the spatial context of points both inside and among local regions.
    % In the point feature expansion, we propose a hierarchical learnable folding strategy to facilitate the feature propagation from sparse to dense in the feature space.
    \item To constrain the distribution of generated points without the ground-truth dense point sets, we introduce a novel self-projection optimization, which interactively projects the generated points onto the underlying object surface along the projection direction. % of each point
    % \item Experimental results demonstrate that the performances of SPU-Net are competitive with state-of-the-art upsampling methods in both qualitative and quantitative measures.
\end{itemize}

To evaluate the performances of SPU-Net, we adopt four widely used metrics to compare with the state-of-the-art methods under a variety of synthetic and real-scanned datasets. 
Experimental results show that our self-supervised method achieves good performance, comparable to the supervised methods in both qualitative and quantitative comparisons.
\section{Related Work}
\noindent\textbf{Traditional point cloud upsampling methods.}
Traditional methods have tried various optimization strategies to generate the upsampled point clouds without using deep learning models.
For example, Alexa \textit{et al.} \cite{alexa2003computing} upsampled the point set by computing the Voronoi diagram and adding points at the vertices of this diagram.
Afterward, the locally optimal projection (LOP) operator \cite{lipman2007parameterization,huang2009consolidation} has been proved to be effective for point resampling and surface reconstruction based on $L_1$ median, especially for point sets with noise and outliers.
Furthermore, Huang \textit{et al.} \cite{huang2013edge} introduced a progressive strategy for edge-aware point set resampling.
To fill large holes and complete missing regions, Dpoints \cite{wu2015deep} developed a new point representation.
In general, all these methods are not in a data-driven manner, which heavily relies on shape priors, such as normal estimation and smooth surface assumption.

\begin{figure*}[htp]
    \centering
    \includegraphics[width=18cm]{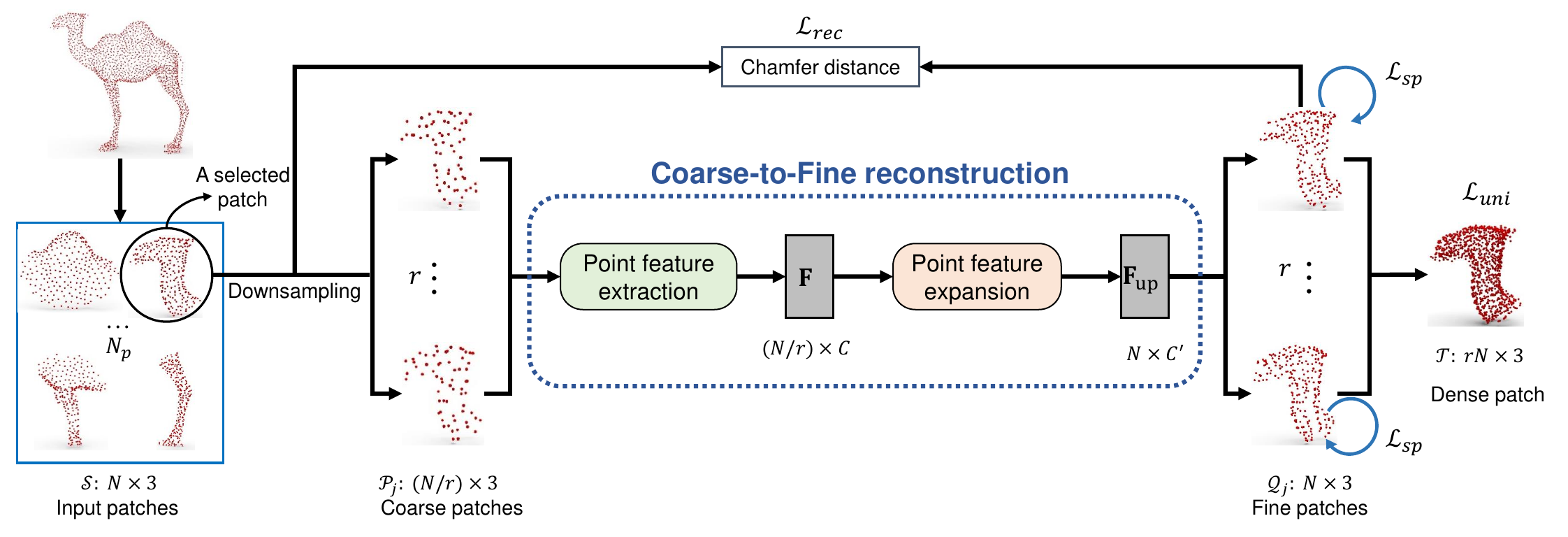}
    \caption{The architecture of SPU-Net. Given an input patch with $N$ points, we first downsample the input patch into $r$ coarse patches $\mathcal{P}_j$, each with $N/r$ points, where $r$ is the upsampling rate.
    And then, we generate a fine patch $\mathcal{Q}_j$ for each coarse patch $\mathcal{P}_j$ in a coarse-to-fine reconstruction framework, which consists of point feature extraction and expansion.
    Finally, we obtain the dense target patch $\mathcal{T}$ by aggregating all the fine patches.
    In addition, $C$ and $C^{'}$ are the number of feature channels that are 480 and 128, respectively, in the implementation; $\mathcal{L}_{rec}$, $\mathcal{L}_{uni}$ and $\mathcal{L}_{sp}$ denote the reconstruction, uniform, and self-projection terms, respectively.}
    \label{fig:main_frame}
\end{figure*}
\noindent\textbf{Deep learning based point cloud upsampling methods.}
In recent years, deep neural networks have achieved outperforming performances in various point cloud processing tasks, including shape classification \cite{liu2019relation,liu2019point2sequence,wen2020point2spatialcapsule,liu2021fine}, object detection \cite{Qi_2020_CVPR,shi2020points}, semantic scene segmentation \cite{hu2020randla,shi2020spsequencenet,wen2020cf}, point cloud reconstruction \cite{ma2020neural,liu2019l2g,han2020reconstructing} and point cloud completion \cite{Wen_2020_CVPR,wang2020cascaded,wen2021pmp,wen2021cycle4completion,Xiang_2021_ICCV}.
In the field of point cloud upsampling, Yu \textit{et al.} \cite{yu2018pu} as a pioneer, first proposed a deep neural network PU-Net to upsample point set, which works on patches by learning multi-level per-point features and expanding the point set via multi-branch convolutions.
Later, they designed another edge-aware point cloud upsampling network named EC-Net \cite{yu2018ec} to achieve point expansion by minimizing the point-to-edge distances.
Wang \textit{et al.} \cite{yifan2019patch} presented a multi-step progressive upsampling network to maintain the patch details further.
Li \textit{et al.} \cite{li2019pu} proposed a GAN-based framework to generate high-quality upsampled point sets.
Recently, Qian \textit{et al.} \cite{qian2020pugeo} incorporated discrete differential geometry to guide the generation of points in the point cloud upsampling.
Li \textit{et al.} \cite{li2021point} proposed a two-step point upsampling strategy, including dense point cloud generating and point spatial refining.
Existing methods can already generate high-quality upsampled point clouds under synthetic datasets with the supervision of dense point sets.
However, these supervised methods require the ground-truth dense point clouds as the supervision information and are not suitable for the real-scanned data.
Liu \textit{et al.} \cite{liu2019l2g} proposed a deep neural network named L2G-AE in representation learning, which can be applied for unsupervised point cloud upsampling by reconstructing overlapped local regions.
However, L2G-AE concentrated on the capturing of global shape information via local-to-global reconstruction, which limits the network in capturing inherent upsampling patterns and generating high-quality upsampled point sets.
To fully explore the spatial patterns inside sparse point sets, we present SPU-Net to upsample from downsampled patches in a coarse-to-fine reconstruction framework, enabling us to generate high-quality upsampled point clouds in a self-supervised manner.

\noindent\textbf{Unsupervied point cloud analysis methods.}
% Most existing methods rely on target labels to build the mapping from input point clouds to the label space.
Recently, several methods have investigated unsupervised or self-supervised learning strategies for point cloud analysis \cite{sauder2019self,sharma2020self,eckart2021self}.
The auto-encoder, such as FoldingNet \cite{yang2018foldingnet} and L2G-AE \cite{liu2019l2g}, is a widely used framework for unsupervised point cloud learning, which adopts the input point cloud itself as the reconstruction target.
During the self-reconstruction process, the corresponding point features are obtained, which can be applied in various applications, such as shape classification \cite{yang2018foldingnet}, semantic segmentation \cite{han2019multi,hassani2019unsupervised}, and shape generation \cite{han2019multi}.
Based on the encoder-decoder architecture, some works conducted self-supervised information as the optimization target, such as coordinate transformation \cite{gao2020graphter}, deformation reconstruction \cite{achituve2020self}, and part partition \cite{zhang2019unsupervised}.
In addition, the generative adversarial network \cite{achlioptas2018learning} was also applied to distinguish the generated point set or the real point set.
In this work, we adopt the generalized encoder-decoder framework to build our self-supervised upsampling network based on local patches.
In order to capture the inherent upsampling patterns, the key issue is how to reproduce the upsampling process without requiring the supervised dense point sets.
To resolve this issue, we propose a coarse-to-fine reconstruction framework to reproduce point cloud upsampling inside local patches.
Specifically, we first downsample the input patch into several subsets, and then the coarse-to-fine reconstruction framework is applied to upsample the subsets.
By repeating the downsampling and upsampling steps, our method can generate dense upsampled point sets. 

\begin{figure*}[htp]
    \centering
    \includegraphics[width=18cm]{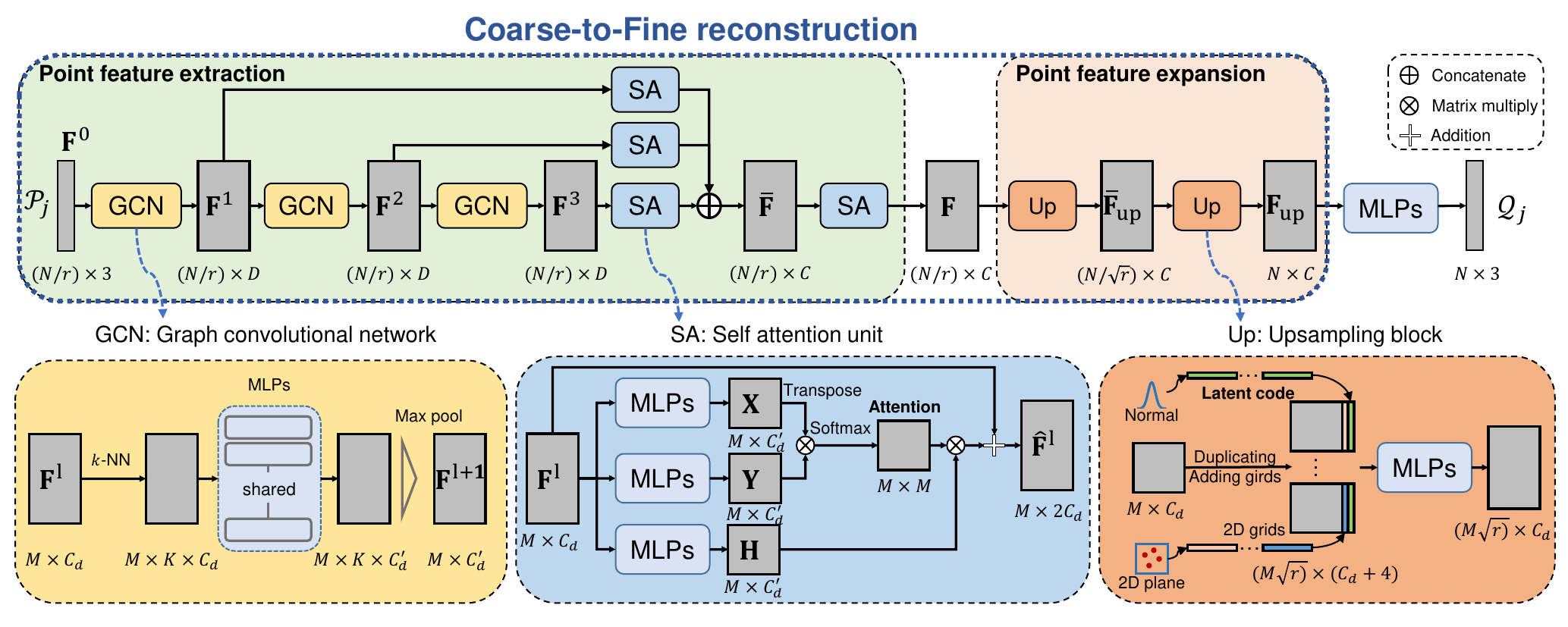}
    \caption{The coarse-to-fine reconstruction framework. Given a coarse patch $\mathcal{P}_j$ with $N/r$ points, we aim to generate the corresponding fine patch $\mathcal{Q}_j$ with $N$ points.
    In this framework, there are two main components: point feature extraction and point feature expansion.
    Here, $D$ and $C$ are the numbers of feature channels that are 64 and 480, respectively, in the implementation.}
    \label{fig:coarse2fine}
\end{figure*}
\section{The SPU-Net Method}
\subsection{Overview}
Given a 3D point cloud, we take the same patch-based approach as PU-Net \cite{yu2018pu} and PU-GAN \cite{li2019pu}.
We first build $N_p$ local patches according to the geodesic distance for the 3D point cloud, as shown in Figure \ref{fig:main_frame}.
For each local patch $\mathcal{S}={\{\bm{s}_i\}}_{i=1}^N$ with $N$ points, our goal is to output a dense and uniform point set $\mathcal{T}={\{\bm{t}_i\}}_{i=1}^{rN}$ with $rN$ points, while keeping the original data distribution, where $\bm{s}_i$, $\bm{t}_i$ are the coordinates of 3D points and $r$ is the upsampling rate. 
Without the supervision of ground-truth dense point sets, we propose a coarse-to-fine reconstruction framework to reproduce the upsampling process inside each local patch.
By upsampling the sparse patch to obtain the fine patch, we are able to capture the inherent upsampling patterns for generating dense patch $\mathcal{T}$ that is uniformly distributed on the underlying object surface.

Figure \ref{fig:main_frame} illustrates the architecture of our SPU-Net.
For an input patch $\mathcal{S}$, we first downsample $\mathcal{S}$ into $r$ different coarse patches $\{\mathcal{P}_1,\cdots, \mathcal{P}_j, \cdots, \mathcal{P}_r\}$, each with $N/r$ points (Section \ref{sec:3.2}).
We then introduce the coarse-to-fine framework to explore the inherent upsampling patterns inside local patches (Section \ref{sec:3.3}), which contains two main components: point feature extraction and point feature expansion. 
Lastly, we present the patch-based training strategy with a joint loss function formed by reconstruction, uniform, self-projection terms (Section \ref{sec:3.4}).

\subsection{Point Set Downsampling} \label{sec:3.2}
Without the supervision of dense point sets, we have to construct some self-supervision to support deep networks to capture the inherent upsampling patterns.
To take advantage of the input patch without supervision, we propose a coarse-to-fine reconstruction framework to generate the upsampled dense patch.
Specifically, as shown in Figure \ref{fig:main_frame}, we first downsample the input patch $\mathcal{S}$ into some coarse patches $\mathcal{P}_j$ to formulate the self-supervision with the input patch itself.
Since the goal of point cloud upsampling is to generate a dense and uniform point set from a sparse input, a relatively uniform initial input is beneficial for this task. 
And the downsampling method largely determines the ability of deep networks to capture the surface distribution of 3D shapes, which should cover the distribution of the entire input patch as uniformly as possible. 
Here, the farthest point sampling (FPS) algorithm as a uniform downsampling strategy is adopted in our method, which can generate more uniform coarse patches than other sampling methods such as random sampling and voxel-grid sampling.
Specifically, we repeat the process of picking out $N/r$ points from the input local patch $\mathcal{S}$ for $r$ times with the FPS algorithm.
Finally, we denote the downsampled coarse patches as $\{\mathcal{P}_1,\cdots, \mathcal{P}_j, \cdots, \mathcal{P}_r\}$.
From the downsampled coarse patches, we can reveal the inherent upsampling process by reconstructing the input patch itself, which makes it possible to infer more dense patches.

\subsection{Coarse-to-Fine Reconstruction} \label{sec:3.3}
In the coarse-to-fine reconstruction framework, there are two main components: point feature extraction and point feature expansion.
In the point feature extraction, we integrate self-attention with the graph convolution network (GCN) to simultaneously capture the spatial context of points both inside and among local regions.
In the point feature expansion, we propose a hierarchical learnable folding strategy to facilitate the feature propagation from sparse to dense in the feature space.

\noindent\textbf{Point feature extraction.}
To capture the context information from discrete points, it is important to extract the spatial correlation of points inside local regions.
The graph convolutional network (GCN) \cite{qi2017pointnet++,wang2019dgcnn,liu2019relation,liu2019point2sequence} has been widely applied to capture the context information inside local regions in existing methods.
However, these methods often ignore capturing the correlation among local regions.
To simultaneously extract the context information both inside and among local regions, we propose a point feature extraction module, which integrates self-attention units with GCNs, as shown in Figure \ref{fig:coarse2fine}.
% Given a point set $\mathcal{P}_j$ as input, we first build a local region centered at each point with $k$-Nearest-Neighbors ($k$-NN) algorithm.
% In a GCN manner, several subsequent Multi-Layer-Perceptrons (MLPs) are employed to extract point features inside each local region and then a max-pooling operation are applied to aggregate local region features.
% Existing methods, such as DGCNN \cite{wang2019dgcnn} and RS-CNN \cite{liu2019relation}, ofen focus on capturing the context information inside local regions while ignoring the correlation among local regions.

Given a coarse patch $\mathcal{P}_j$ with the size of $N/r \times 3$ as input, three GCNs are first employed to capture the local contexts inside local regions by building a local graph around each point $\bm{p}^j_i$. 
We introduce a hierarchical feature extraction strategy with multiple semantic levels.
Suppose the input feature map of a GCN at level $l \in \{0, 1, 2, 3\}$ is $\bm{F}^l = \{ \bm{f}_i^l \}_{i=1}^M$ with the size of $M \times C_d$, where $\bm{f}_i^l$ is the $i$-th point feature in $\bm{F}^l$.
In particular, the feature map $\bm{F}^{0}$ at level 0 is the raw points from $\mathcal{P}_j$.
To calculate the point feature $\bm{f}^{l+1}_i$, we first dynamically build a local region $\mathcal{N}^l_i$ with $K$ neighbors around each point feature $\bm{f}^l_i$ with the $k$-NN algorithm.
And then, we formulate the propagation of point feature $\bm{f}^l_i$ ($l \in \{0,1,2\}$) as
\begin{equation}
    \bm{f}^{l+1}_i = \mathop{max}_{\bm{f}^l_j \in \mathcal{N}^l_i} \{\sigma( \bm{h_{\theta}} (\bm{f}^l_j - \bm{f}^l_i)\}.
\end{equation}
Here, $(\bm{f}^l_j - \bm{f}^l_i)$ can be regarded as the edge from point $\bm{f}^l_j$ to center point $\bm{f}^l_i$, $\bm{h_{\theta}}$ indicates the learnable parameters in multi-layer-perceptrons (MLPs), $\sigma$ is a non-linear layer, such as ReLU \cite{nair2010rectified}, and $\mathop{max}$ is a max-pool operation.
The max-pool layer is applied to aggregate point features in the local region $\mathcal{N}^l_i$.
% With three stacked GCNs,  multi-level point features $\{\bm{f}^1_i,\bm{f}^2_i,\bm{f}^3_i\}$ of $s_i \in \mathcal{P}_j$ are captured, which contains the local structures of local regions.

% Relying on local structures cannot cover the overall geometric information of the input point set, the relationship between local regions also plays a important role in point feature extraction.
Following previous works \cite{liu2019l2g,li2019pu}, we integrate self-attention units to capture the correlation between local regions.
As shown in Figure \ref{fig:coarse2fine}, the self-attention unit cooperates with the GCN to capture the detailed spatial context information inside local patches.
Suppose that the input feature map is $\bm{F}^l$ with the size of $M \times C_d$.
Three MLPs are used to embed $\bm{F}^l$ into different feature spaces, $\bm{X}$, $\bm{Y}$ and $\bm{H}$, respectively.
In particular, $\bm{X}$ and $\bm{Y}$ are applied to calculate the attention values with simple matrix multiplication, and the updated feature map $\bm{\hat{F}}^l$ is calculated as 
\begin{equation}
    \bm{\hat{F}^l} = \bm{F}^l + softmax(\bm{Y} \bm{X}^\top)\bm{H}.
\end{equation}

After multiple self-attention units, the correlation among local regions is captured at different semantic levels.
We then aggregate the multi-level point features by concatenation as 
\begin{equation}
    \bm{\bar{F}} = MLP(\bm{F}^1 \oplus \bm{\hat{F}}^1 \oplus \bm{\hat{F}}^2 \oplus \bm{\hat{F}}^3).
\end{equation}

To further explore the correlation in the aggregated feature $\bm{\bar{F}}$, we add another self-attention unit to obtain the final point feature $\bm{F}$.

\noindent\textbf{Point feature expansion.}
The target of point feature expansion is to construct the mapping from current points to more points, which is widely used in some point-wise applications, such as semantic segmentation \cite{qi2017pointnet++}, point cloud reconstruction \cite{achlioptas2018learning} and point cloud upsampling \cite{li2019pu}.
In general, current point feature expansion methods can be roughly categorized into the interpolating-based method \cite{qi2017pointnet++}, folding-based method \cite{yang2018foldingnet} and reshaping-based method \cite{achlioptas2018learning}. 
Existing point interpolating-based methods often use the point interpolation relationship to guide the expansion of point features.
However, in many scenarios, the interpolation relationship between point sets is usually unknown.
In addition, reshaping-based methods usually first expand the feature dimensions with deep networks, such as MLPs or fully-connected (FC) layers, and then generate the target point features via a simple reshaping operation.
In recent years, the folding-based method has been developed, which first duplicates point features and then concatenates 2D grids to guide point feature expansion.

Compared with other feature expansion methods, the folding-based method is more flexible and has achieved satisfactory performances in various applications \cite{yang2018foldingnet,Wen_2020_CVPR}.
However, previous fixed 2D grids are not adaptive to various feature distributions.
To resolve this problem, we propose novel learnable 2D grids as the latent code to cooperate with fixed 2D grids in guiding the point feature expansion.
As shown in Figure \ref{fig:coarse2fine}, given an input feature map with the size of $M \times C_d$, we first duplicate the point features and then concatenate two kinds of grids.
In particular, the latent code is initialized from a standard normal distribution and can be optimized in the network training process.
Moreover, in order to smooth the point feature expansion process, we introduce the hierarchical folding strategy.
By using two upsampling blocks, we hierarchically obtain the upsampled point features $\bm{\bar{F}}_{up}$ and $\bm{F}_{up}$ with an upsampling rate $\sqrt{r}$.
With subsequent MLPs, the point features $\bm{F}_{up}$ are applied to reconstruct the fine patches.

\subsection{Loss Function} \label{sec:3.4}
\noindent \textbf{Joint loss.}
In our SPU-Net, we optimize the upsampled point set with a joint loss $\mathcal{L}_{joint}$, consisting of reconstruction term $\mathcal{L}_{rec}$, uniform term $\mathcal{L}_{uni}$ and self-projection term $\mathcal{L}_{sp}$.
Overall, we train our SPU-Net by minimizing the joint loss function in an end-to-end manner as
\begin{equation}
    \label{eq:joint}
    \mathcal{L}_{joint} = \alpha \mathcal{L}_{rec} + \beta \mathcal{L}_{uni} + \gamma \mathcal{L}_{sp},
\end{equation}
where $\alpha$, $\beta$ and $\gamma$ are hyper-parameters in training.

\noindent\textbf{Reconstruction term.}
For an input local patch $\mathcal{S}$, we first downsample the input patch into some coarse patches, and then we repeat upsampling from each one of the coarse patches to explore the inherent upsampling patterns in a self-supervised manner.
Assume that $\mathcal{Q}_j = \{ \bm{q}^j_i\}_{i=1}^N$ is the upsampled fine patch from the $j$-th coarse patch $\mathcal{P}_j$.
Thus, we formulate the reconstruction term using chamfer distance (CD) for $\mathcal{Q}_j$ as
\begin{equation}
\begin{aligned}
   \mathcal{L}_{rec} = d_{CD}(\mathcal{S}, \mathcal{Q}_j) &=\sum_{i=1}^N{\frac{1}{|\mathcal{S}|} \sum_{\bm{s}_i \in \mathcal{S}}{\min_{\bm{q}^j_i \in \mathcal{Q}_j}||\bm{s}_i - \bm{q}^j_i||_2}} \\ 
   &+\sum_{i=1}^N{\frac{1}{|\mathcal{Q}^j|} \sum_{\bm{q}^j_i \in \mathcal{Q}_j}{\min_{\bm{s}_i \in \mathcal{S}}||\bm{s}_i - \bm{q}^j_i||_2}}.
\end{aligned}
\end{equation}

\noindent \textbf{Uniform term.}
The reconstruction term encourages the upsampled fine patch $\mathcal{Q}_j$ to fit the input patch $\mathcal{S}$.
However, the target dense patch $\mathcal{T} = \{\mathcal{Q}_1, \cdots, \mathcal{Q}_j, \cdots, \mathcal{Q}_r\}$ should be uniformly distributed on the underlying object surface.
Similar to \cite{li2019pu,yu2018pu}, we introduce a uniform term to improve the uniformity of the final target patch $\mathcal{T}$ with size of $rN \times 3$.
%  the uniform loss evaluates the uniformity of patch $\mathcal{T}$, which contains point number constraint and point distance constraint.
Specifically, we first use the farthest sampling to pick $M$ seed points in $\mathcal{T}$ and apply the ball query of radius $r_d$ to build a local region (denoted as $\mathcal{T}_j, j = 1, \cdots, M$) in $\mathcal{T}$ around each seed. Thus, the number of points roughly relies on the small disk of area $\pi r_d^2$ on the underlying surface.
In addition, the patch $\mathcal{T}$ is normalized in a unit sphere with an area of $\pi 1^2$.
So, the expected number of points $\hat{n}$ in $\mathcal{T}_j$ is $rNr^2_d$. 
Suppose the local region $\mathcal{T}_j$ satisfies the regular hexagonal distribution and the distance $d_{j,k}$ is for $k$-th point in $\mathcal{T}_j$.
So, the expected point-to-neighbor distance $\hat{d}$ should be roughly $\sqrt{2\pi r^2_d /(|\mathcal{T}_j|\sqrt{3})}$.
To measure the deviation of $|\mathcal{T}_j|$ from $\hat{n}$ and $d_{j,k}$ from $\hat{d}$, we calculate the uniform term with chi-square as 
\begin{equation}
\begin{aligned}
    \mathcal{L}_{uni}{(\mathcal{T}_j)} &=\sum_{j=1}^M U_{number}{(\mathcal{T}_j)} \cdot U_{distance}(\mathcal{T}_j) \\
    &=\sum_{j=1}^M \frac{(|\mathcal{T}_j| - \hat{n})^2}{\hat{n}} \cdot \sum_{k=1}^{|\mathcal{T}_j|}{\frac{(d_{j,k}-\hat{d})^2}{\hat{d}}}.
\end{aligned}
\end{equation}

\begin{figure}[tp]
\centering 
\includegraphics[width=9cm]{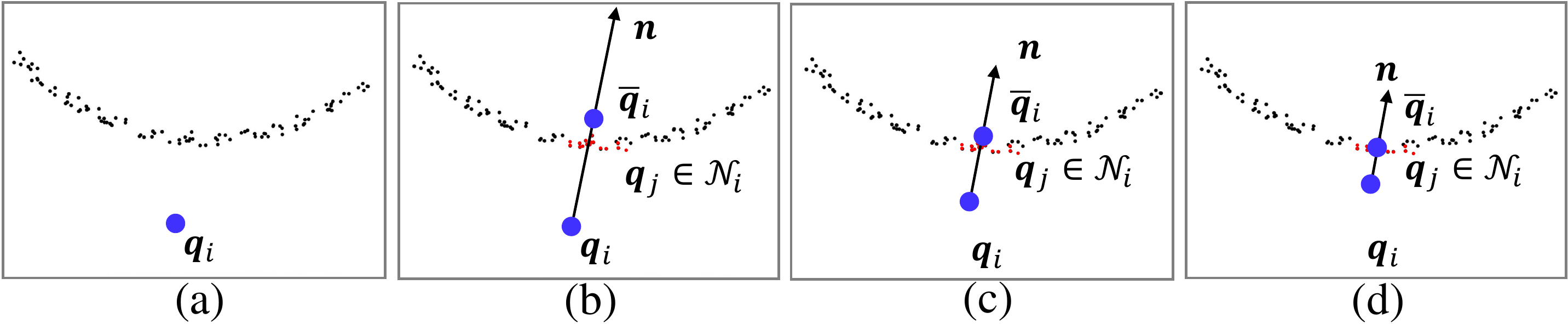}
\caption{
    The illustration of the optimization effect of the self-projection loss term $\mathcal{L}_{sp}$. During the training, the noisy point $\bm{q}_i$ can be gradually optimized to the underlying object surface by minimizing the self-projection term $\mathcal{L}_{sp}$, where $\mathcal{L}_{sp}$ is calculated by measuring the local distribution of noisy point $\bm{q}_i$, neighbor point $\bm{q}_j$ and local center $\bar{\bm{q}}_i$ in the local region $\mathcal{N}_i$.
    % The illustration of self-projection optimization. During training, the noisy point $\bm{q}_i$ is gradually pulled onto the underlying object surface, where the projection direction $\bm{n}$ is calculated by measuring the local distribution of $\bm{q}_i$ and local center $\bar{\bm{q}}_i$ in the local region $\mathcal{N}_i$.
    } %  %大图名称
\label{fig:projection}  %图片引用标记
\end{figure}
\noindent \textbf{Self-projection term.}
The reconstruction term and uniform term optimize the geometric shape and surface distribution of the generated upsampled point sets, respectively.
However, due to the irregular nature of the point cloud and network bias in the generation of dense point clouds from sparse ones, it is inevitable to bring some noisy points in the generated upsampled point sets.
To resolve this problem, we propose a novel self-projection loss function to constrain the generated points to lie roughly on the underlying object surface.
Inspired by the traditional directed projection (DP) algorithm \cite{Liu06-LSP}, each point in the generated set can be projected to the underlying object surface by directed projection.
Differently, our SPU-Net is a learning-based method, which constrains the generated point by loss functions.
Therefore, our self-projection loss term aims to characterize the local distribution of points and penalize the noise distribution, where the projection process of points is implicit.
% So, we should first find a projection direction for each point in the generated point set.
% One option is to project each point to the distribution of the input patch.
Considering the distribution differences between input sparse patches and generated dense patches, we formulate the function inside local regions in a self-projection manner. 
Suppose the generated fine patch is $\mathcal{Q} = \{ \bm{q}_i\}_{i=1}^N$ with a size of $N \times 3$.
We first build a local region $\mathcal{N}_i$ around the $i$-th point of $\mathcal{Q}$ with the $k$-nearest-neighbors algorithm.
Then, we calculate the center point $\bar{\bm{q}}_i = \frac{1}{|\mathcal{N}_i|} \sum_{\bm{q}_j \in \mathcal{N}_i}{(\bm{q}_j)}$ of the local region $\mathcal{N}_i$ to be the projection target of the current point $\bm{q}_j$.
Therefore, the self-projection optimization is formulated with chi-square as
\begin{equation}
    \mathcal{L}_{sp} = \frac{1}{|\mathcal{Q}|\cdot |\mathcal{N}_j|}\sum_{i=1}^{|\mathcal{Q}|}{\sum_{\bm{q}_j \in \mathcal{N}_i}                   {\frac{|(\bm{q}_i - \bm{q}_j)^2 - (\bar{\bm{q}}_i- \bm{q}_j)^2|}{1+(\bm{q}_i - \bm{q}_j)^2}}}.
\end{equation}
Here, $\frac{1}{1+(\bm{q}_i - \bm{q}_j)^2}$ is the weight of each point in the local region $\mathcal{N}_j$.
As shown in Figure \ref{fig:projection}, $(\bm{q}_i - \bm{q}_j)^2$ and $(\bar{\bm{q}}_i - \bm{q}_j)^2$ indicate the real distribution and the target distribution of the local region, respectively. 
\begin{figure}[tp]
    \centering
    \includegraphics[width=8.5cm]{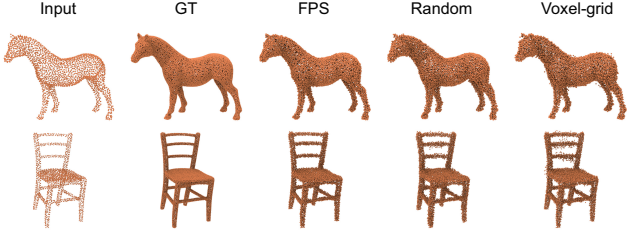}
    \caption{The visualization of upsampling results in different downsampling strategies under the PU-GAN dataset.}
    \label{fig:sampling}
\end{figure}
\begin{table}[tp]
\centering
\caption{The effects of the point downsampling strategies.}
\label{table:sampling}
\begin{tabular}{ccccc}\hline
Metric &FPS &RS &MS &VS\\ \hline
CD ($10^{-3}$) &\textbf{0.38} &0.52 &0.55 &0.66 \\ \hline
HD ($10^{-3}$) &\textbf{2.24} &4.19 &4.27  &7.20 \\ \hline
P2F ($10^{-3}$) &\textbf{5.87} &7.57 &7.22 &9.79 \\ \hline
UNI ($10^{-3}$) &\textbf{8.94} &14.89 &9.9 &14.60 \\ \hline
\end{tabular}
\end{table}
\section{Experiments}
In this section, we first investigate how some key hyper-parameters affect the point cloud upsampling performance of our SPU-Net.
Then, we evaluate SPU-Net by comparing it with state-of-the-art point cloud upsampling methods qualitatively and quantitatively.
In addition, we also show some upsampling visualizations under KITTI \cite{geiger2013vision}, and ScanNet \cite{dai2017scannet}, which illustrates the advantage of our SPU-Net trained without supervision.
And we engage in an ablation study of network components in SPU-Net to evaluate the effectiveness of different components.
Finally, we conduct a computational complexity analysis of our approach, where the SPU-Net is comparable with other compared methods.

\subsection{Datasets and Network Configurations}
For a fair comparison, we do both quantitative and qualitative comparisons with state-of-the-art methods under the dataset from PU-GAN \cite{li2019pu}, which contains 147 3D models that are formed with 120 objects in the training dataset and 27 objects in the test set.
And we follow the same patch-based training strategy as in PU-Net \cite{yu2018pu}, and PU-GAN \cite{li2019pu}.
By default, we crop 24 patches for each training model and set the number of patch points $N = 256$, the upsampling rate $r = 4$.
Moreover, as for the uniform term, the number of seed points $M$ is 50, and we cropped the same set of $\mathcal{T}_j$ with radius $r_d = \sqrt{p}$ for each $p \in  \{0.4 \%, 0.6 \%, 0.8 \%, 1.0 \%, 1.2 \% \}$ as in \cite{li2019pu}.
% To avoid overfitting in training, we augment the network inputs by random rotation, scaling, and point perturbation with Gaussian noisy. 
We train the network for 200 epochs with the Adam algorithm \cite{kingma2014adam}. 
Moreover, we set the learning rate as 0.0001, and we gradually reduce both rates by a decay rate of 0.7 per 50k iterations until $10^{-6}$. 
The batch size is 24, and $\alpha$, $\beta$, and $\gamma$ are empirically set as 100, 10, and 0.01 in Eq. (\ref{eq:joint}), respectively. 
We implemented our network with TensorFlow and trained it on Nvidia 2,080 Ti GPU.
    
To evaluate the performance of point cloud upsampling, we employ the same four evaluation metrics as PU-GAN \cite{li2019pu}, including uniformity (UNI), point-to-surface (P2F), Chamfer distance (CD), and Hausdorff distance (HD). For quantitative evaluation, we use Poisson disk sampling to sample 2,048 points as training data for each 3D object and 8,192 points as the ground truth upsampled point cloud in the testing phase.

\begin{figure}[tp]
\centering 
\includegraphics[width=8.5cm]{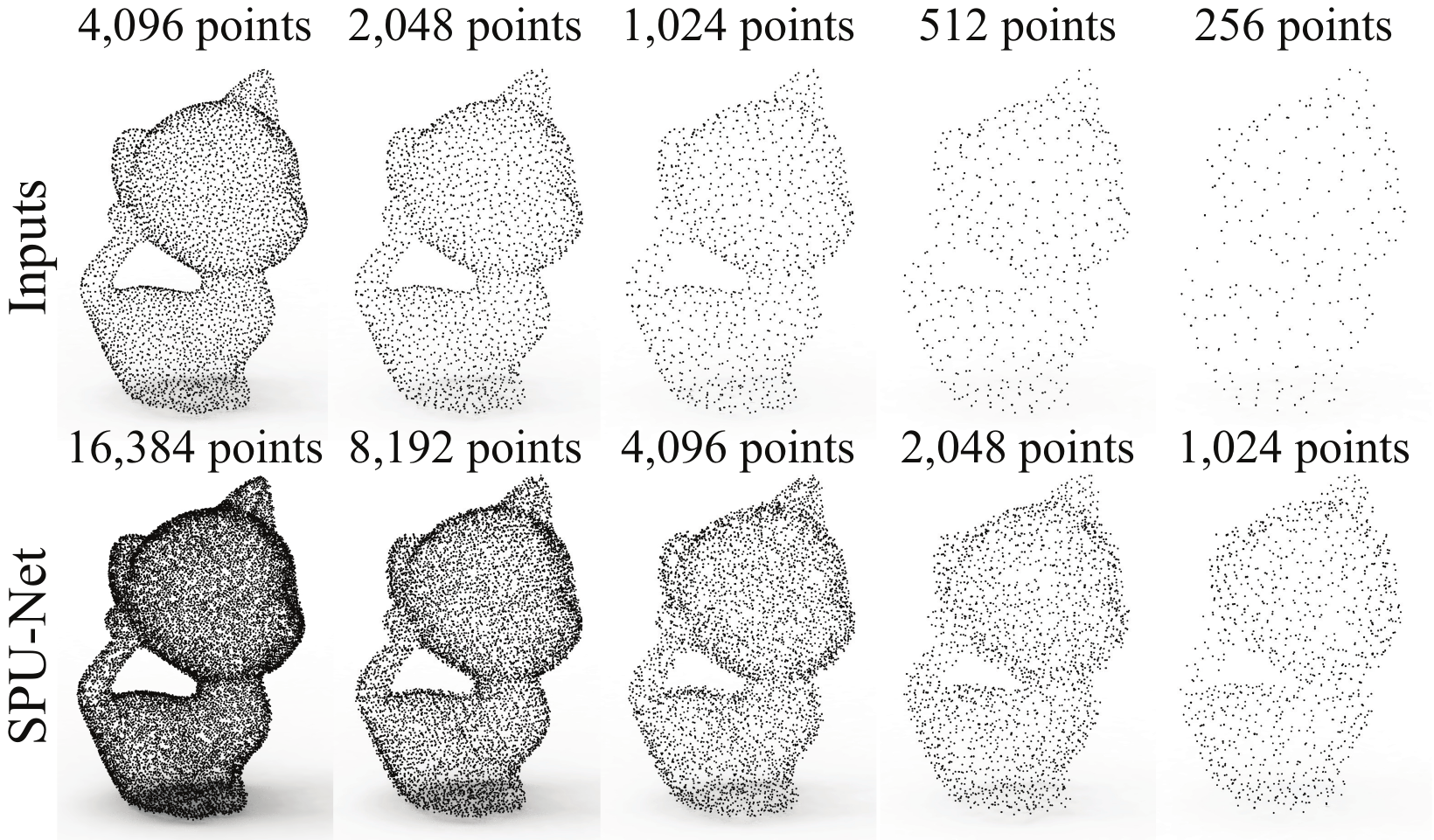}
\caption{The upsampling results with varying input sizes from 256 points to 4,096 points.} %  %大图名称
\label{fig:sparsity}  %图片引用标记
\end{figure}
\begin{table}[tp]
\centering
\caption{The effects of the number of region points $K$.}
\label{table:pfe_k}
% \resizebox{8.5cm}{8mm}{
\begin{tabular}{cccccc}\hline
Metric &$K=5$ &$10$&$15$ &$20$ \\ \hline
CD ($10^{-3}$) &0.45 &\textbf{0.38} &0.39 &0.41 \\ \hline
HD ($10^{-3}$) &3.03 &\textbf{2.24} &2.51 &2.48 \\ \hline
P2F ($10^{-3}$) &5.91 &\textbf{5.87} &6.04 &6.12 \\ \hline
UNI ($10^{-3}$) &12.47 &\textbf{8.94} &9.65 &9.90 \\ \hline
\end{tabular}
\end{table}

\begin{table*}[tp]
\centering
\caption{Quantitative comparisons with the state-of-the-arts.}
\label{table:compare}
% \resizebox{8.5cm}{14mm}{
\begin{tabular}{l|c|ccccc|c|c|c}\hline \hline
\multirow{2}{*}{Methods} &
\multirow{2}{*}{Supervised?} &
\multicolumn{5}{c|}{Uniformity for different $p$ $(10^{-3})$} &
\multirow{2}{*}{\makecell[c]{P2F \\ $(10^{-3})$}}&
\multirow{2}{*}{\makecell[c]{CD \\ $(10^{-3})$}}&
\multirow{2}{*}{\makecell[c]{HD \\ $(10^{-3})$}} \\
% second row
\cline{3-7}
&\multicolumn{1}{c|}{}
&\multicolumn{1}{c}{$0.4\%$}
&\multicolumn{1}{c}{$0.6\%$} 
&\multicolumn{1}{c}{$0.8\%$} 
&\multicolumn{1}{c}{$1.0\%$} 
&\multicolumn{1}{c|}{$1.2\%$} 
&\multicolumn{1}{c|}{} 
&\multicolumn{1}{c|}{} 
&\multicolumn{1}{c}{}
\\ \hline \hline
PU-Net \cite{yu2018pu}      &Yes &29.74 &31.33 &33.86 &36.94 &40.43 &6.84 &0.72 &8.94  \\ \hline
MPU \cite{yifan2019patch}   &Yes &7.51 &7.41 &8.35 &9.62 &11.13 &3.96 &0.49 &6.11  \\ \hline
PU-GAN \cite{li2019pu}      &Yes &3.38 &3.49 &3.44 &3.91 &4.64 &2.33 &0.28 &4.64  \\ \hline
Dis-PU \cite{li2021point}   &Yes   &\textbf{2.47} &\textbf{1.93} &\textbf{2.21} &\textbf{2.75} &\textbf{3.48} &\textbf{2.01} &\textbf{0.22} &\textbf{2.83} \\  \hline \hline
EAR \cite{EAR2013}          &No &16.84 &20.27 &23.98 &26.15 &29.18 &5.82 &0.52 &7.37 \\ \hline
L2G-AE \cite{liu2019l2g}    &No &24.61 &34.61 &44.86 &55.31 &64.94 &39.37 &6.31 &63.23 \\ \hline
Ours (Train2Test)      &No &\textbf{4.53} &\textbf{4.82} &\textbf{5.68} &\textbf{6.69} &\textbf{7.95} &5.97 &0.38 &2.24  \\ \hline 
Ours (All2Test)        &No &4.71 &5.02 &5.91 &7.03 &8.50 &\textbf{5.79} &\textbf{0.37} &2.55  \\ \hline
Ours (Test2Test)       &No &4.82 &5.14 &5.86 &6.88 &8.13 &6.85 &0.41 &\textbf{2.18}  \\ \hline \hline
\end{tabular}
\end{table*}
\subsection{Parameters}
All the experiments in parameter comparisons are evaluated under the dataset from PU-GAN, where Chamfer distance (CD), Hausdorff distance (HD), point-to-surface (P2F), and uniformity (UNI, $p=1.2\%$) are adopted as evaluation metrics.
For each point cloud with 2,048 points, we sample $N_p = 24$ seed points with FPS and build local patches according to geodesic distance.
We initialize the network hyper-parameters, as depicted in the network configurations.
Given an input patch, we first downsample this patch into some coarse patches, which should cover the input patch with a uniform distribution in the downsampling process.
Thus, the sampling methods influence the neural network to capture inherent upsampling patterns.
In Table \ref{table:sampling}, we report the results of three different sampling strategies, including farthest point sampling (FPS), random sampling (RS), mixing of these two sampling strategies (MS) and voxel-grid sampling (VS).
The results show that FPS can generate relatively uniform distributions to cover input patches and facilitate point cloud upsampling.
The qualitative upsampling results with different downsampling strategies are also illustrated in Figure \ref{fig:sampling}, where the FPS can promote better upsampling results than other strategies.

Then, we investigate the impact of the number of local points $K$ in GCNs of the point feature extraction.
Specifically, we range the number of local region points $K$ from 5 to 20.
Table \ref{table:pfe_k} illustrates the results of different $K$.
The best performance achieves at $K=10$, which can effectively capture the local region context inside local regions.

Moreover, we also evaluate the robustness of our SPU-Net under different sparsity input point clouds. 
Figure \ref{fig:sparsity} shows the upsampling point sets with varying sizes of input points from 256 to 4,096.
Our method is stable even for input with only 256 points.
% More hyper-parameter comparisons are given in the \textbf{Supplemental Material}.

\begin{figure*}[tp]
\centering 
\includegraphics[width=\textwidth]{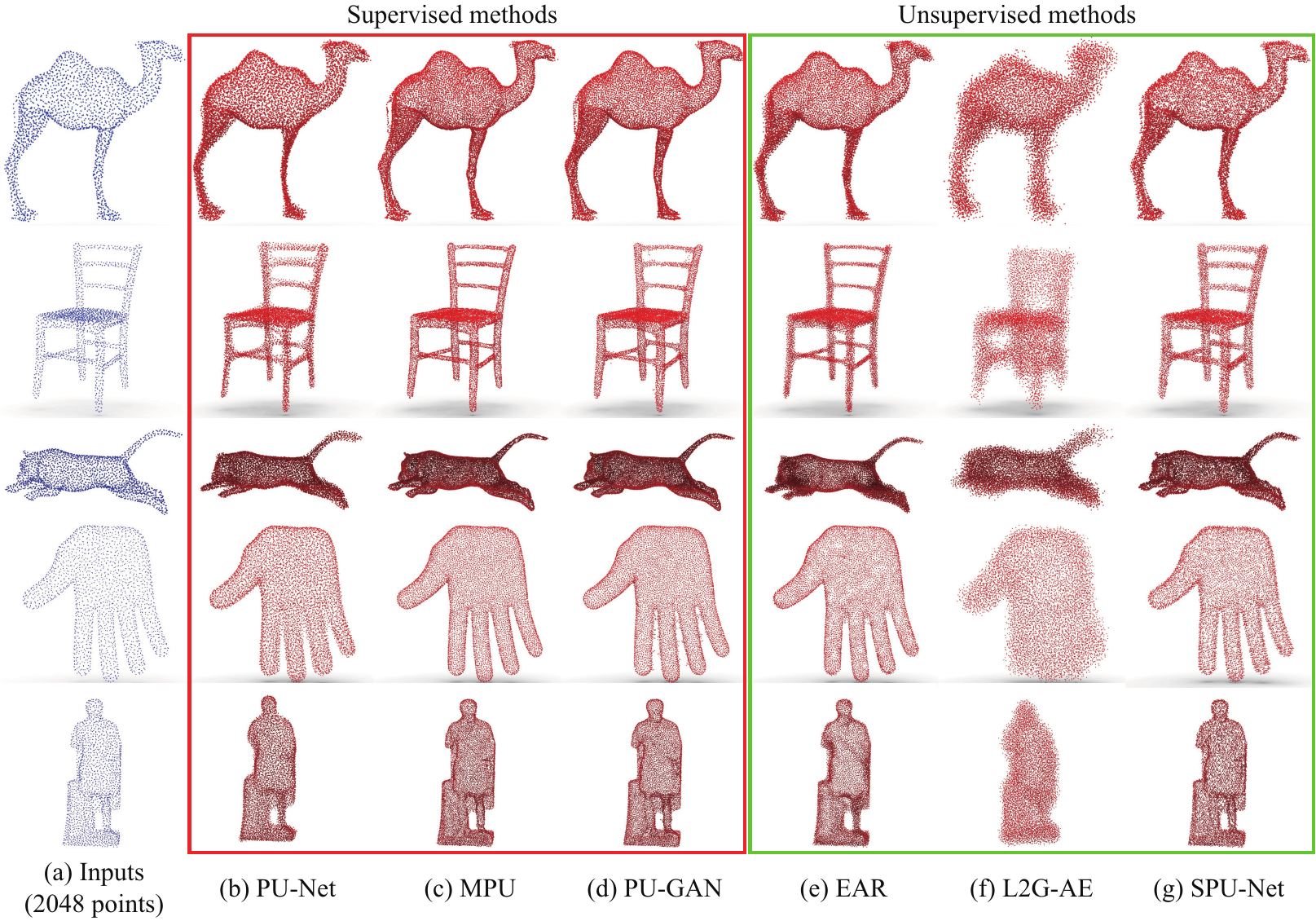}
\caption{The visualization results of upsampling from 2,048 points to 8,192 points under the dataset provided by \cite{li2019pu}. With only a sparse point cloud as input, our SPU-Net can also achieve comparable results as supervised methods, where the neural network and loss function simultaneously act on the generated point clouds.}
\label{fig:upsampling}  %图片引用标记
\end{figure*}

\begin{table}[tp]
\centering
\caption{The effects of the number of patches and the number of points inside each patch.}
\label{table:patch_points}
\resizebox{8.5cm}{8mm}{
\begin{tabular}{cccccccccc}\hline
Metric &$N=128$ &$256$&$384$ &$512$ &$N_p = 12$ &$18$ &$30$ &$36$\\ \hline
CD ($10^{-3}$) &0.45 &0.38 &0.44 &0.41 &0.59 &0.43 &0.40 &0.39\\ \hline
HD ($10^{-3}$) &3.26 &2.24 &2.76 &2.50 &4.53 &2.77 &2.37 &2.38\\ \hline
P2F ($10^{-3}$) &5.84 &5.87 &6.55 &7.04 &6.32 &6.23 &6.31 &6.25\\ \hline
UNI ($10^{-3}$) &16.89 &8.94 &9.31 &8.76 &22.00 &11.70 &8.82 &8.28\\ \hline
\end{tabular}}
\end{table}
\begin{table}[tp]
\centering
\caption{The effects of the number of local points $k_g$ in calculating geodesic distance.}
\label{table:geo_k}
% \resizebox{8.5cm}{8mm}{
\begin{tabular}{cccccc}\hline
Metric &$k_p=3$ &$4$&$5$ &$6$ \\ \hline
CD ($10^{-3}$) &0.39 &0.39 &0.38 &0.47 \\ \hline
HD ($10^{-3}$) &2.25 &2.30 &2.24 &3.18 \\ \hline
P2F ($10^{-3}$) &6.10 &6.09 &5.87 &5.93 \\ \hline
UNI ($10^{-3}$) &9.11 &9.30 &8.94 &13.64 \\ \hline
\end{tabular}
\end{table}
Subsequently, we explore the number of sampled points $N$ in each input patch, which influences the distribution of local patches in point clouds. 
In the experiment, we keep the settings of our network as depicted in the network configuration section and modify the number of patch points $N$ from 128 to 512.
The results are shown in Table \ref{table:patch_points}, where the upsampling metrics on the benchmark have a tendency to rise first and then fall. 
In general, the network reaches the best performance when $N=256$.
Then, we keep the point number $N=256$ and investigate the effect of the number of patches $N_p$ for each 3D shape.
We changed the number of patches $N_p$ from 12 to 36.
From the results shown in Table \ref{table:patch_points}, the three metrics slightly fluctuate when $N_p$ goes from 18 to 36, which shows that these patch number settings can cover the entire input point cloud well.
On the contrary, the patch number $N_p=12$ can not cover all the information of point clouds. 

Therefore, we employ the number of patches $N_p = 24$ and the number of patch points $N = 256$ as the setting of our network in the following experiments. Then, as shown in Table \ref{table:geo_k}, we show the effect of the number of local points $k_g$ in calculating geodesic distance.
We ranged the local point $k_g$ from 3 to 6.
The above results suggest that $k_g = 5$ is more suitable for our network and achieves the best performances.

Finally, to investigate the ratio of loss functions, we range the ratio $\alpha$, $\beta$ and $\gamma$ of loss functions $L_{rec}$, $L_{uni}$ and $L_{pro}$ as shown in Table \ref{table:loss}.
The initialized loss ratios are $\alpha=100$, $\beta=10$ and $\gamma=0.01$.
From the experimental results, there is a trade-off between the loss function ratio and the evaluation metric. 
Therefore, we choose the initialization loss function ratio as the final loss setting.

\begin{table}[tp]
\centering
\caption{The effects of the loss function ratio $\alpha,\beta$ and $\gamma$.}
\label{table:loss}
\resizebox{8.5cm}{8mm}{
\begin{tabular}{ccccccccc}\hline
Metric &$\alpha=50$ &$100$ &$200$ &$\beta=5$ &$20$ &$\gamma=0.005$ &$0.02$\\ \hline
CD ($10^{-3}$) &0.49 &0.38 &0.39 &0.39 &0.40 &0.37 &0.38  \\ \hline
HD ($10^{-3}$) &3.36 &2.24 &2.64 &2.36 &2.26 &2.32 &2.35 \\ \hline
P2F($10^{-3}$) &9.16 &5.87 &5.55 &5.43 &6.55 &6.04 &6.13 \\ \hline
UNI ($10^{-3}$) &13.81 &8.94 &9.46 &9.53 &9.21 & 8.12 &10.83 \\ \hline
\end{tabular}}
\end{table}

\begin{figure}[tp]
    \centering
    \includegraphics[width=8cm]{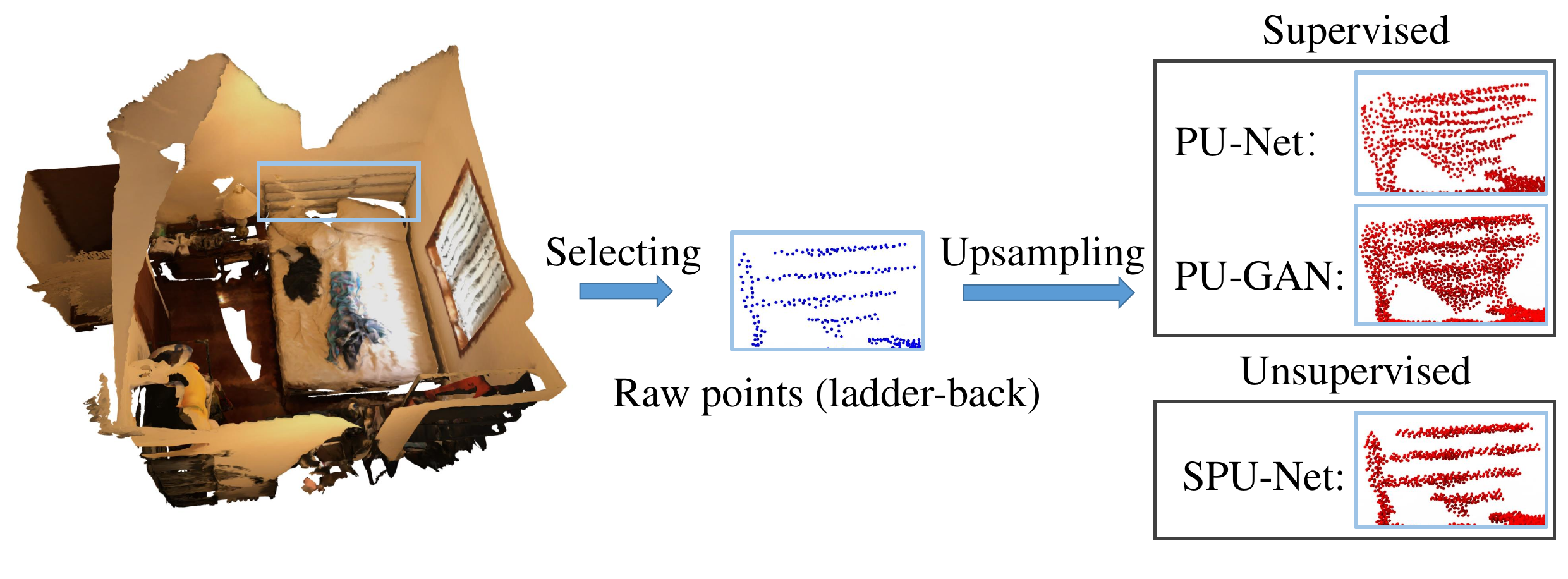}
    \caption{The upsampling results of a real-scanned object (bed) from the ScanNet dataset \cite{dai2017scannet}. The supervised methods (e.g., PU-Net \cite{yu2018pu} and PU-GAN \cite{li2019pu}) destroy the surface distribution of raw points (the ladder-back of the bed). In contrast, our SPU-Net keeps the origin point distribution.}
    \label{fig:idea}
\end{figure}
\begin{figure}[tp]
    \centering
    \includegraphics[width=8.5cm]{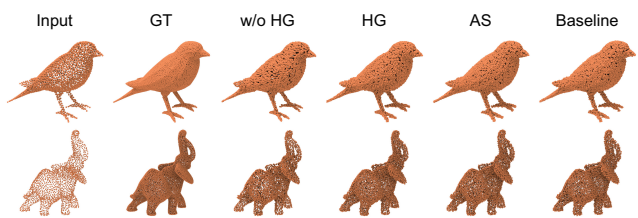}
    \caption{The visualization of different settings in the folding operation.}
    \label{fig:grids}
\end{figure}

\begin{figure*}[htp]
    \centering
    \includegraphics[width=15cm]{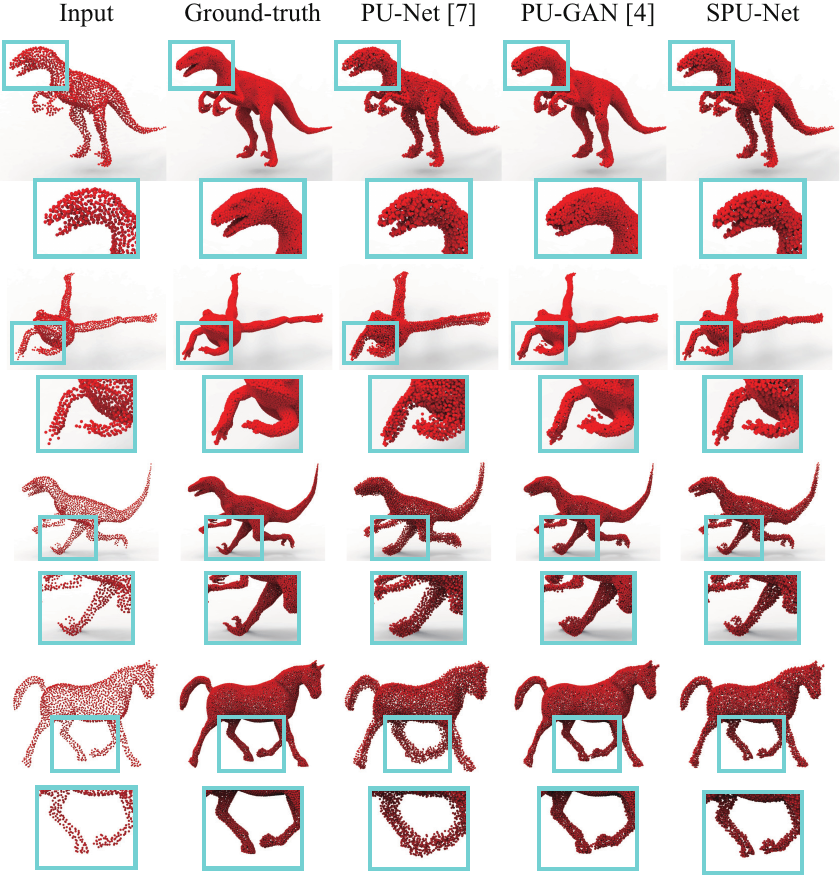}
    \caption{The upsampling results under the ISDB dataset \cite{gal2007pose}. For each input sparse point cloud, we generate a 8,192 dense point set from 2,408 input points.}
    \label{fig:isdb}
\end{figure*}

\begin{figure*}[htp]
    \centering
    \includegraphics[width=14cm]{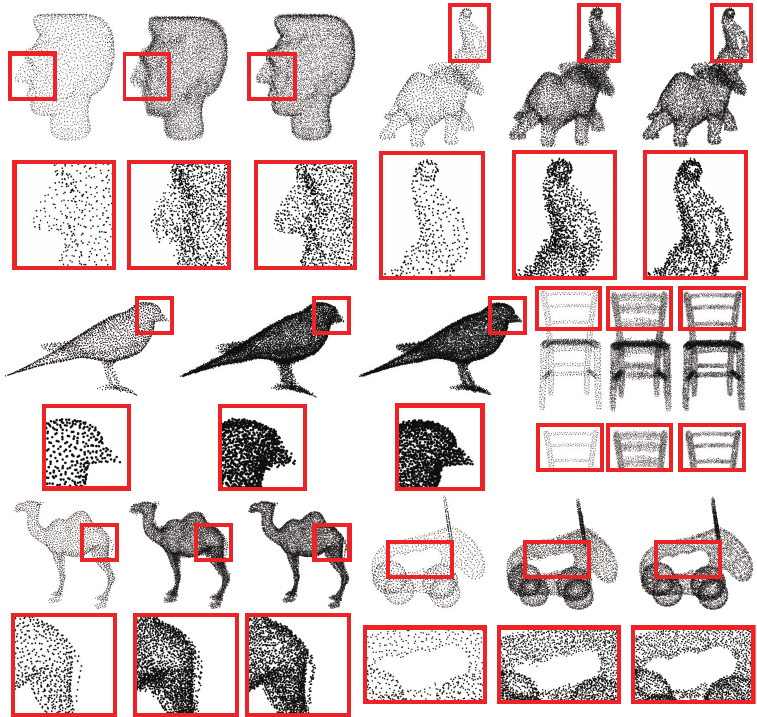}
    \caption{The effect of the self-projection term. There are six test 3D point clouds from the dataset of PU-GAN \cite{li2019pu}.
For each point cloud, we show the sparse input point cloud (left), the upsampled point cloud without the self-projection term (middle), and the upsampled point cloud with the self-projection term (right).
The red boxes show the details brought about by the self-projection term.}
    \label{fig:sp}
\end{figure*}

\subsection{Results on Synthetic Dataset}
We compare our SPU-Net with several state-of-the-art upsampling methods both quantitatively and qualitatively, including EAR \cite{EAR2013}, PU-Net \cite{yu2018pu}, MPU \cite{yifan2019patch}, PU-GAN \cite{li2019pu}, Dis-PU \cite{li2021point} and L2G-AE \cite{liu2019l2g}.
For EAR, we use its demo code and generate the best results by fine-tuning the associated parameters.
Note that, for the recent work PUGeoNet \cite{qian2020pugeo}, we cannot report the comparison results due to the unavailable code and trained model so far.
In addition, as for recent Dis-PU \cite{li2021point}, we report the quantitative results tested with public code and trained model.
In Table \ref{table:compare}, our SPU-Net achieves comparable results with existing supervised upsampling networks, such as PU-Net \cite{yu2018pu}, and outperforms another unsupervised method L2G-AE \cite{liu2019l2g}. 
All evaluation metrics are the same as the ones employed in PU-GAN, where the uniformity is evaluated with varying scales $p$.
And we report multiple testing results of our SPU-Net under different training datasets, including only the training dataset (Train2Test), only the test dataset (Test2Test), and the entire dataset (All2Test).
In addition, we also show the qualitative comparisons in Figure \ref{fig:upsampling}, which shows that our SPU-Net can generate upsampled point sets with more details.

To further evaluate our SPU-Net, we also visualize some upsampling results from ISDB \cite{gal2007pose}. 
ISDB contains about 104 articulated models of animals and humans, some of that have slight differences between models, which is challenging for the point cloud upsampling task.
For each shape, we first sample 2,048 points from the 3D mesh with Poisson Disk Sampling \cite{corsini2012efficient}.
We aim to generate a dense point set with 8,192 points for each sparse input set with 2,048 points.
Here, we also evaluate some supervised methods with the trained models under the dataset from PU-GAN, including PU-Net \cite{yu2018pu} and PU-GAN \cite{li2019pu}.
For our SPU-Net, we directly train our network under the ISDB dataset.
From the results in Figure \ref{fig:isdb}, we find that our SPU-Net can well preserve the detailed information of the input sparse point clouds.

\begin{figure*}[htp]
    \centering
    \includegraphics[width=14cm]{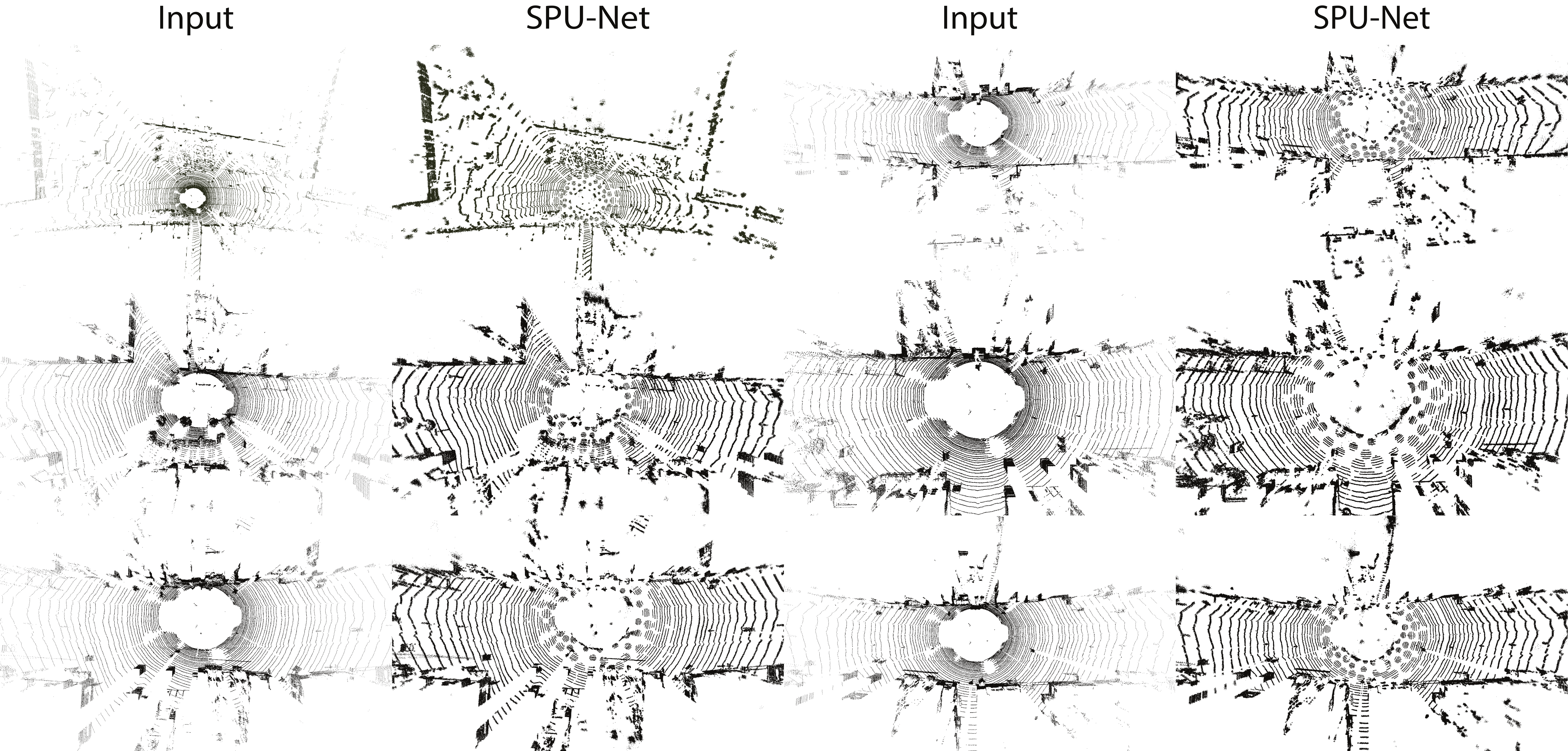}
    \caption{The upsampling results ($\times 4$) under KITTI dataset \cite{geiger2013vision}. We divide the real-scanned scenes into small patches with 256 points for each in training.}
    \label{fig:kitti}
\end{figure*}

\begin{figure*}[htp]
    \centering
    \includegraphics[width=14cm]{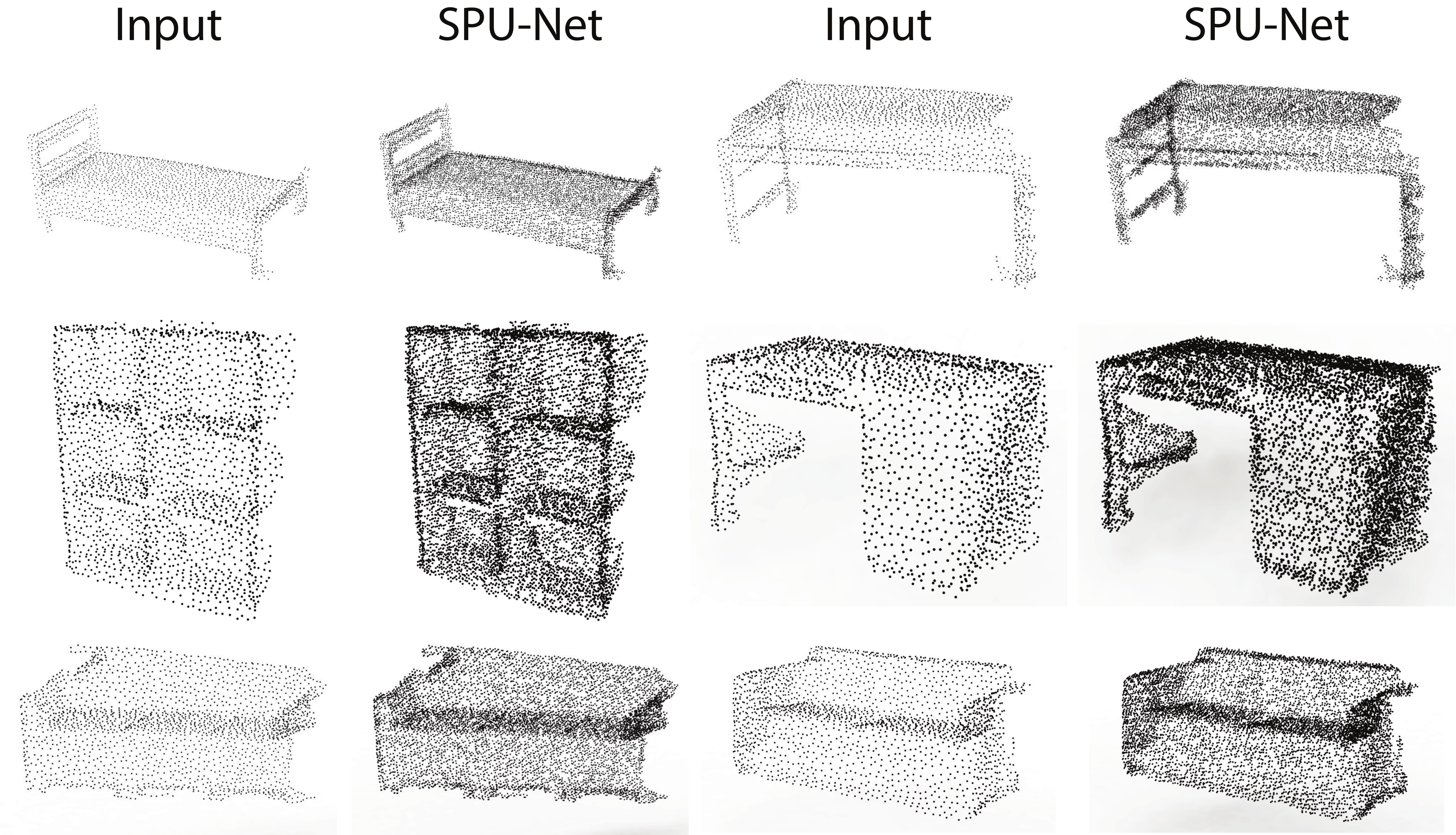}
    \caption{The upsampling results ($\times 4$) under ScanNet dataset \cite{dai2017scannet}. We pick out objects from indoor scenes and divide them into small pathches with 256 points for each for training.}
    \label{fig:scannet}
\end{figure*}
\subsection{Results on Real-scanned Dataset}
To evaluate the performance of SPU-Net, we directly train the network under the real-scanned datasets, including KITTI \cite{geiger2013vision} and ScanNet \cite{dai2017scannet}.
As for preparing training data, we first apply FPS to sample a certain number of seed points and then use KNN to build local patches around these points.
Different from supervised methods, our method can be directly trained under real-scanned data, and SPU-Net can preserve the details of raw data as shown in Figure \ref{fig:idea}.
Figures \ref{fig:kitti} and \ref{fig:scannet} show our upsampling results on real-scanned point sets. 
From the results, our SPU-Net can generate dense and uniform upsampled point sets from sparse ones.
In particular, all results under the real-scanned data are directly trained on the raw points without ground-truth dense point sets.
Therefore, our SPU-Net can expand the training data from synthetic data to the real-scanned data.

\begin{table}[tp]
\centering
\caption{The effect of some key components in the SPU-Net.}
\label{table:ablation}
\resizebox{8.5cm}{8mm}{
\begin{tabular}{cccccccccc}\hline
Metric          &Baseline            &NSA   &NHG            &NLG    &AS   &NRT        &NUT            &NST        &GP\\ \hline
CD ($10^{-3}$)  &\textbf{0.35}  &0.44   &0.68           &0.36   &0.42   &4.68       &0.38           &0.65       &0.41 \\ \hline
HD ($10^{-3}$)  &\textbf{2.20}  &2.99   &5.11           &2.36   &2.86   &34.58      &2.55           &6.24       &2.53 \\ \hline
P2F ($10^{-3}$) &5.24           &6.45   &\textbf{3.95}  &5.47   &5.17   &40.44      &4.49           &9.01       &5.93\\ \hline
UNI ($10^{-3}$) &\textbf{9.65}  &12.23  &30.48          &9.97   &11.15   &802.64     &13.49          &19.38      &10.35 \\ \hline
\end{tabular}}
\end{table}

\subsection{Ablation Study}
To evaluate the components in SPU-Net, including the coarse-to-fine framework (i.e., removing the self-attention unit) and loss function terms,  we remove each of them and generate upsampling results for testing models.
Specifically, we remove the self-attention unit (No self-attention, NSA), the learnable grids (No learnable grid, NLG), the hierarchical grids (No hierarchical grid, NHG), the reconstruction term (No reconstruction term, NRT), the uniform loss (No uniform term, NUT) and the self-projection term (No self-projection term, NST), respectively.
All above ablation studies are compared with the full network pipeline (Baseline) and we also show the results of adding supervision at the middle folding layer (AS).
Specifically, we apply a single-direction CD loss to constrain the folding process in the AS setting.
In addition, we also replace the whole coarse-to-fine framework with the generator of PU-GAN as a new baseline (Generator of PU-GAN, GP) to show the effectiveness of our coarse-to-fine framework.
The results in Table \ref{table:ablation} suggest that all key components play an important role in improving the performance of our SPU-Net, where the network components and the loss function work together to capture the inherent upsampling patterns.
In particular, we show the visualization results in Figure \ref{fig:sp}, which demonstrate the effectiveness of the self-projection loss function in optimizing noisy points to the underlying object surface itself.
To intuitively show the impact of the folding settings, some qualitative results are displayed in Figure \ref{fig:grids}, including without hierarchical grids (w/o HG), only hierarchical grids (HG), adding supervision in the middle folding layer (AS) and our base model (Baseline).
The results show that the hierarchical grids are important for constraining the point distribution of point upsampling. And some middle supervision might be helpful for optimizing the final distribution of points.
\begin{figure}[htp]
    \centering
    \includegraphics[width=8.5cm]{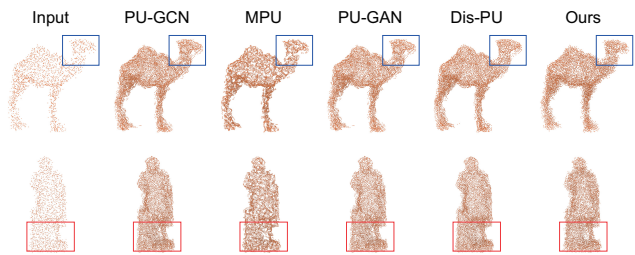}
    \caption{The comparison of different methods under PU-GAN data with noise level of 2\%. The colored boxes represent some local details of the point cloud upsampling.}
    \label{fig:noise}
\end{figure}

\begin{figure*}[htp]
    \centering
    \includegraphics[width=15cm]{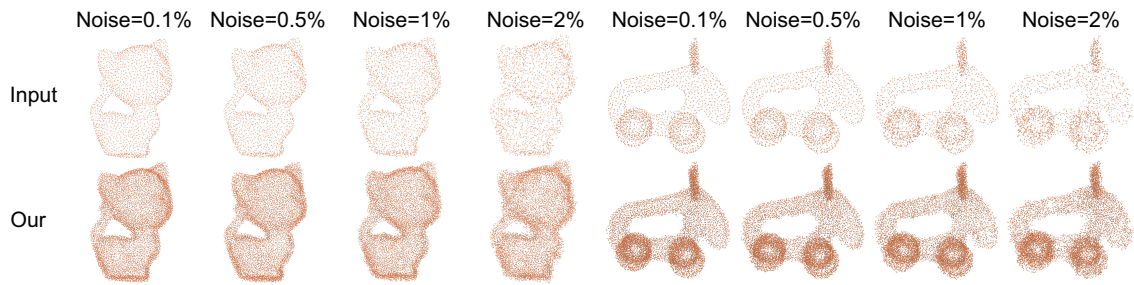}
    \caption{The result of our SPU-Net under different noise levels of 0.1\%, 0.5\%, 1\%, and 2\%. Our SPU-Net is relatively robust to the input }
    \label{fig:selfnoise}
\end{figure*}

\noindent\textbf{Noise effect.} The upsampling results under different levels of noise are revealed in Figure \ref{fig:selfnoise}. In our SPU-Net, the proposed network learns to infer the dense points in a self-supervised manner, which is a challenging task. In order to constrain the distribution of upsampled points, we introduce a joint loss function, including reconstruction term, uniform term and self-projection term. Benefitted from the self-projection term, our SPU-Net is robust to generate dense points under different noise levels of 0.1\%, 0.5\%, 1\% and 2\%.

As shown in Figure \ref{fig:noise}, we also visually compare the results with different methods under PU-GAN data with noise level 2\%. Benefit from the joint loss function and coarse-to-fine upsampling strategy, our SPU-Net can explore upsampling patterns inside input sparse patches. From the results, our SPU-Net can keep the details of the input sparse point clouds, which demonstrates our good generalization ability to noisy point clouds.

\begin{figure*}[htp]
    \centering
    \includegraphics[width=15cm]{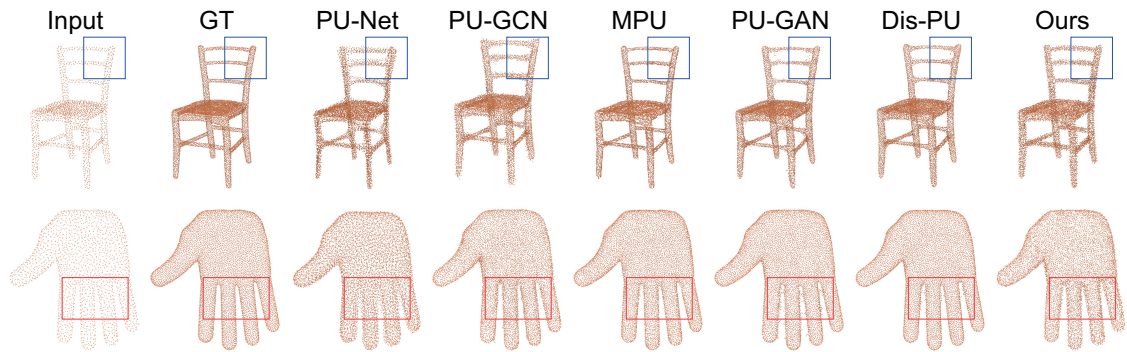}
    \caption{Some comparisons of different upsampling methods under PU-GAN dataset.}
    \label{fig:special}
\end{figure*}
\noindent\textbf{Local distribution comparisons.} We compare our SPU-Net with existing supervised point cloud upsampling methods under the PU-GAN dataset in Figure \ref{fig:special}. There are some situations that existing supervised methods cannot solve very well. In some local regions, such as the ladder-back of the chair and the fingers of the hand, our SPU-Net can obtain a relatively clean distribution in the coarse-to-fine point upsampling.

\subsection{Computational Complexity}
In this section, we do a computational complexity analysis of our SPU-Net by comparing it with some state-of-the-art upsampling methods.
In Table \ref{table:complexity}, we adopt the GFLOPs, training parameters, and forward time as the evaluation metrics.
The compared methods include PU-Net \cite{yu2018pu}, MPU \cite{yifan2019patch}, PU-GAN \cite{li2019pu} and L2G-AE \cite{liu2019l2g}.
We reproduce their results using their released code, where the statistical analysis is conducted using the built-in functions in TensorFlow.
Before the evaluation, we initialize the batch size of all methods to 1.
The comparison in Table \ref{table:complexity} demonstrates that our method is efficient in terms of computational performance.
\begin{table}[tp]
\centering
\caption{The analysis of computational complexity.}
\label{table:complexity}
\resizebox{8.5cm}{10mm}{
\begin{tabular}{lccc}\hline
Methods &GFLOPs &Training parameters (Mb) &Forward time (ms) \\ \hline
PU-Net \cite{yu2018pu} &15.03 &0.81&135.88 \\ \hline
MPU \cite{yifan2019patch} &27.52 &\textbf{0.30}&368.81\\ \hline
PU-GAN \cite{li2019pu}&5.67 &0.68&146.16\\ \hline
L2G-AE \cite{liu2019l2g}&15.43 &100.18&185.10\\ \hline
\textbf{Ours (SPU-Net)} &\textbf{1.86} &0.68&\textbf{130.92}\\ \hline
\end{tabular}}
\end{table}
\section{Conclusion}
In this paper, we propose a novel self-supervised point cloud upsampling method to generate dense and uniform point set from sparse inputs without the supervision of ground-truth dense point clouds.
Our coarse-to-fine reconstruction framework effectively facilitates point cloud upsampling by point feature extraction and point feature expansion.
In addition, our self-projection optimization successfully projects noisy points onto the underlying object surface itself, which greatly improves the quality of point cloud upsampling in an unsupervised manner. 
Our experimental results demonstrate that our method can achieve good performance on both synthetic and real-scanned datasets, even comparable results to the state-of-the-art supervised method.

\bibliographystyle{IEEEtran}
\bibliography{IEEEabrv,reference}

\begin{thebibliography}{10}
\providecommand{\url}[1]{#1}
\csname url@samestyle\endcsname
\providecommand{\newblock}{\relax}
\providecommand{\bibinfo}[2]{#2}
\providecommand{\BIBentrySTDinterwordspacing}{\spaceskip=0pt\relax}
\providecommand{\BIBentryALTinterwordstretchfactor}{4}
\providecommand{\BIBentryALTinterwordspacing}{\spaceskip=\fontdimen2\font plus
\BIBentryALTinterwordstretchfactor\fontdimen3\font minus
  \fontdimen4\font\relax}
\providecommand{\BIBforeignlanguage}[2]{{%
\expandafter\ifx\csname l@#1\endcsname\relax
\typeout{** WARNING: IEEEtran.bst: No hyphenation pattern has been}%
\typeout{** loaded for the language `#1'. Using the pattern for}%
\typeout{** the default language instead.}%
\else
\language=\csname l@#1\endcsname
\fi
#2}}
\providecommand{\BIBdecl}{\relax}
\BIBdecl

\bibitem{yu2018pu}
L.~Yu, X.~Li, C.-W. Fu, D.~Cohen-Or, and P.-A. Heng, ``{PU-Net: Point Cloud
  Upsampling Network},'' in \emph{{Proceedings of the IEEE Conference on
  Computer Vision and Pattern Recognition}}, 2018, pp. 2790--2799.

\bibitem{yifan2019patch}
W.~Yifan, S.~Wu, H.~Huang, D.~Cohen-Or, and O.~Sorkine-Hornung, ``{Patch-Based
  Progressive 3D Point Set Upsampling},'' in \emph{{Proceedings of the IEEE
  Conference on Computer Vision and Pattern Recognition}}, 2019, pp.
  5958--5967.

\bibitem{li2019pu}
R.~Li, X.~Li, C.-W. Fu, D.~Cohen-Or, and P.-A. Heng, ``{PU-GAN: A Point Cloud
  Upsampling Adversarial Network},'' in \emph{{Proceedings of the IEEE
  International Conference on Computer Vision}}, 2019, pp. 7203--7212.

\bibitem{qian2020pugeo}
Y.~Qian, J.~Hou, S.~Kwong, and Y.~He, ``{PUGeo-Net: A Geometry-centric Network
  for 3D Point Cloud Upsampling},'' in \emph{Proceedings of the European
  Conference on Computer Vision}, 2020, pp. 752--769.

\bibitem{li2021point}
R.~Li, X.~Li, P.-A. Heng, and C.-W. Fu, ``{Point Cloud Upsampling via
  Disentangled Refinement},'' \emph{Proceedings of the IEEE Conference on
  Computer Vision and Pattern Recognition}, 2021.

\bibitem{chang2015shapenet}
A.~X. Chang, T.~Funkhouser, L.~Guibas, P.~Hanrahan, Q.~Huang, Z.~Li,
  S.~Savarese, M.~Savva, S.~Song, H.~Su \emph{et~al.}, ``{ShapeNet: An
  Information-Rich 3D Model Repository},'' \emph{arXiv preprint
  arXiv:1512.03012}, 2015.

\bibitem{Visionair-dataset}
``{VisionAir},'' \url{http://www.infra-visionair.eu/}, accessed: 2021-06-28.

\bibitem{dai2017scannet}
A.~Dai, A.~X. Chang, M.~Savva, M.~Halber, T.~Funkhouser, and M.~Nie{\ss}ner,
  ``{ScanNet: Richly-Annotated 3D Reconstructions of Indoor Scenes},'' in
  \emph{{Proceedings of the IEEE International Conference on Computer Vision}},
  2017.

\bibitem{geiger2013vision}
A.~Geiger, P.~Lenz, C.~Stiller, and R.~Urtasun, ``{Vision meets robotics: The
  KITTI dataset},'' \emph{The International Journal of Robotics Research},
  vol.~32, no.~11, pp. 1231--1237, 2013.

\bibitem{yuan2018unsupervised}
Y.~Yuan, S.~Liu, J.~Zhang, Y.~Zhang, C.~Dong, and L.~Lin, ``Unsupervised image
  super-resolution using cycle-in-cycle generative adversarial networks,'' in
  \emph{Proceedings of the IEEE Conference on Computer Vision and Pattern
  Recognition Workshops}, 2018, pp. 701--710.

\bibitem{shaham2019singan}
T.~R. Shaham, T.~Dekel, and T.~Michaeli, ``{SinGAN: Learning a Generative Model
  from a Single Natural Image},'' in \emph{{Proceedings of the IEEE
  International Conference on Computer Vision}}, 2019, pp. 4570--4580.

\bibitem{liu2019l2g}
X.~Liu, Z.~Han, X.~Wen, Y.-S. Liu, and M.~Zwicker, ``{L2G Auto-Encoder:
  Understanding Point Clouds by Local-to-Global Reconstruction with
  Hierarchical Self-Attention},'' in \emph{{Proceedings of the ACM
  International Conference on Multimedia}}, 2019, pp. 989--997.

\bibitem{alexa2003computing}
M.~Alexa, J.~Behr, D.~Cohen-Or, S.~Fleishman, D.~Levin, and C.~T. Silva,
  ``{Computing and Rendering Point Set Surfaces},'' \emph{{IEEE Transactions on
  Visualization and Computer Graphics}}, vol.~9, no.~1, pp. 3--15, 2003.

\bibitem{lipman2007parameterization}
Y.~Lipman, D.~Cohen-Or, D.~Levin, and H.~Tal-Ezer, ``{Parameterization-Free
  Projection for Geometry Reconstruction},'' \emph{{ACM Transactions on
  Graphics}}, vol.~26, no.~3, pp. 22--es, 2007.

\bibitem{huang2009consolidation}
H.~Huang, D.~Li, H.~Zhang, U.~Ascher, and D.~Cohen-Or, ``{Consolidation of
  Unorganized Point Clouds for Surface Reconstruction},'' \emph{{ACM
  Transactions on Graphics}}, vol.~28, no.~5, pp. 1--7, 2009.

\bibitem{huang2013edge}
H.~Huang, S.~Wu, M.~Gong, D.~Cohen-Or, U.~Ascher, and H.~Zhang, ``{Edge-Aware
  Point Set Resampling},'' \emph{ACM Transactions on Graphics}, vol.~32, no.~1,
  pp. 1--12, 2013.

\bibitem{wu2015deep}
S.~Wu, H.~Huang, M.~Gong, M.~Zwicker, and D.~Cohen-Or, ``{Deep Points
  Consolidation},'' \emph{{ACM Transactions on Graphics}}, vol.~34, no.~6, pp.
  1--13, 2015.

\bibitem{liu2019relation}
Y.~Liu, B.~Fan, S.~Xiang, and C.~Pan, ``{Relation-Shape Convolutional Neural
  Network for Point Cloud Analysis},'' in \emph{{Proceedings of the IEEE
  Conference on Computer Vision and Pattern Recognition}}, 2019, pp.
  8895--8904.

\bibitem{liu2019point2sequence}
X.~Liu, Z.~Han, Y.-S. Liu, and M.~Zwicker, ``{Point2Sequence: Learning the
  Shape Representation of 3D Point Clouds with an Attention-Based Sequence to
  Sequence Network},'' in \emph{Proceedings of the AAAI Conference on
  Artificial Intelligence}, vol.~33, 2019, pp. 8778--8785.

\bibitem{wen2020point2spatialcapsule}
X.~Wen, Z.~Han, X.~Liu, and Y.-S. Liu, ``{Point2SpatialCapsule: Aggregating
  Features and Spatial Relationships of Local Regions on Point Clouds using
  Spatial-Aware Capsules},'' \emph{IEEE Transactions on Image Processing},
  vol.~29, pp. 8855--8869, 2020.

\bibitem{liu2021fine}
X.~Liu, Z.~Han, Y.-S. Liu, and M.~Zwicker, ``{Fine-Grained 3D Shape
  Classification with Hierarchical Part-View Attention},'' \emph{IEEE
  Transactions on Image Processing}, vol.~30, pp. 1744--1758, 2021.

\bibitem{Qi_2020_CVPR}
C.~R. Qi, X.~Chen, O.~Litany, and L.~J. Guibas, ``{ImVoteNet: Boosting 3D
  Object Detection in Point Clouds With Image Votes},'' in \emph{{Proceedings
  of the IEEE Conference on Computer Vision and Pattern Recognition}}, June
  2020.

\bibitem{shi2020points}
S.~Shi, Z.~Wang, J.~Shi, X.~Wang, and H.~Li, ``{From Points to Parts: 3D Object
  Detection from Point Cloud with Part-Aware and Part-Aggregation Network},''
  \emph{IEEE Transactions on Pattern Analysis and Machine Intelligence}, 2020.

\bibitem{hu2020randla}
Q.~Hu, B.~Yang, L.~Xie, S.~Rosa, Y.~Guo, Z.~Wang, N.~Trigoni, and A.~Markham,
  ``{RandLA-Net: Efficient Semantic Segmentation of Large-Scale Point
  Clouds},'' in \emph{{Proceedings of the IEEE/CVF Conference on Computer
  Vision and Pattern Recognition}}, 2020, pp. 11\,108--11\,117.

\bibitem{shi2020spsequencenet}
H.~Shi, G.~Lin, H.~Wang, T.-Y. Hung, and Z.~Wang, ``{SpSequenceNet: Semantic
  Segmentation Network on 4D Point Clouds},'' in \emph{{Proceedings of the
  IEEE/CVF Conference on Computer Vision and Pattern Recognition}}, 2020, pp.
  4574--4583.

\bibitem{wen2020cf}
X.~Wen, Z.~Han, G.~Youk, and Y.-S. Liu, ``{CF-SIS: Semantic-Instance
  Segmentation of 3D Point Clouds by Context Fusion with Self-Attention},'' in
  \emph{Proceedings of the ACM International Conference on Multimedia}, 2020,
  pp. 1661--1669.

\bibitem{ma2020neural}
B.~Ma, Z.~Han, Y.-S. Liu, and M.~Zwicker, ``{Neural-Pull: Learning Signed
  Distance Functions from Point Clouds by Learning to Pull Space onto
  Surfaces},'' in \emph{Proceedings of the International Conference on Machine
  Learning}, 2021.

\bibitem{han2020reconstructing}
Z.~Han, B.~Ma, Y.-S. Liu, and M.~Zwicker, ``{Reconstructing 3D Shapes from
  Multiple Sketches using Direct Shape Optimization},'' \emph{IEEE Transactions
  on Image Processing}, vol.~29, pp. 8721--8734, 2020.

\bibitem{Wen_2020_CVPR}
X.~Wen, T.~Li, Z.~Han, and Y.-S. Liu, ``{Point Cloud Completion by
  Skip-Attention Network With Hierarchical Folding},'' in \emph{Proceedings of
  the IEEE Conference on Computer Vision and Pattern Recognition}, 2020, pp.
  1939--1948.

\bibitem{wang2020cascaded}
X.~Wang, M.~H. Ang~Jr, and G.~H. Lee, ``{Cascaded Refinement Network for Point
  Cloud Completion},'' in \emph{{Proceedings of the IEEE/CVF Conference on
  Computer Vision and Pattern Recognition}}, 2020, pp. 790--799.

\bibitem{wen2021pmp}
X.~Wen, P.~Xiang, Z.~Han, Y.-P. Cao, P.~Wan, W.~Zheng, and Y.-S. Liu,
  ``{PMP-Net: Point Cloud Completion by Learning Multi-Step Point Moving
  Paths},'' in \emph{Proceedings of the IEEE/CVF Conference on Computer Vision
  and Pattern Recognition}, 2021, pp. 7443--7452.

\bibitem{wen2021cycle4completion}
X.~Wen, Z.~Han, Y.-P. Cao, P.~Wan, W.~Zheng, and Y.-S. Liu,
  ``{Cycle4Completion: Unpaired Point Cloud Completion using Cycle
  Transformation with Missing Region Coding},'' in \emph{Proceedings of the
  IEEE/CVF Conference on Computer Vision and Pattern Recognition}, 2021, pp.
  13\,080--13\,089.

\bibitem{Xiang_2021_ICCV}
P.~Xiang, X.~Wen, Y.-S. Liu, Y.-P. Cao, P.~Wan, W.~Zheng, and Z.~Han,
  ``{SnowflakeNet: Point Cloud Completion by Snowflake Point Deconvolution With
  Skip-Transformer},'' in \emph{Proceedings of the IEEE/CVF International
  Conference on Computer Vision}, 2021, pp. 5499--5509.

\bibitem{yu2018ec}
L.~Yu, X.~Li, C.-W. Fu, D.~Cohen-Or, and P.-A. Heng, ``{EC-Net: An Edge-Aware
  Point Set Consolidation Network},'' in \emph{{Proceedings of the European
  Conference on Computer Vision}}, 2018, pp. 386--402.

\bibitem{sauder2019self}
J.~Sauder and B.~Sievers, ``{Self-Supervised Deep Learning on Point Clouds by
  Reconstructing Space},'' in \emph{Proceedings of the Advances in Neural
  Information Processing Systems}, 2019, pp. 12\,962--12\,972.

\bibitem{sharma2020self}
C.~Sharma and M.~Kaul, ``{Self-Supervised Few-Shot Learning on Point Clouds},''
  in \emph{Proceedings of the Advances in Neural Information Processing
  Systems}, 2020, pp. 7212--7221.

\bibitem{eckart2021self}
B.~Eckart, W.~Yuan, C.~Liu, and J.~Kautz, ``{Self-Supervised Learning on 3D
  Point Clouds by Learning Discrete Generative Models},'' in \emph{Proceedings
  of the IEEE/CVF Conference on Computer Vision and Pattern Recognition}, 2021,
  pp. 8248--8257.

\bibitem{yang2018foldingnet}
Y.~Yang, C.~Feng, Y.~Shen, and D.~Tian, ``{FoldingNet: Point Cloud Auto-Encoder
  via Deep Grid Deformation},'' in \emph{{Proceedings of the IEEE Conference on
  Computer Vision and Pattern Recognition}}, 2018, pp. 206--215.

\bibitem{han2019multi}
Z.~Han, X.~Wang, Y.-S. Liu, and M.~Zwicker, ``{Multi-Angle Point Cloud-VAE:
  Unsupervised Feature Learning for 3D Point Clouds from Multiple Angles by
  Joint Self-Reconstruction and Half-to-Half Prediction},'' in
  \emph{{Proceedings of the IEEE International Conference on Computer
  Vision}}.\hskip 1em plus 0.5em minus 0.4em\relax IEEE, 2019, pp.
  10\,441--10\,450.

\bibitem{hassani2019unsupervised}
K.~Hassani and M.~Haley, ``{Unsupervised Multi-Task Feature Learning on Point
  Clouds},'' in \emph{{Proceedings of the IEEE International Conference on
  Computer Vision}}, 2019, pp. 8160--8171.

\bibitem{gao2020graphter}
X.~Gao, W.~Hu, and G.-J. Qi, ``{GraphTER: Unsupervised Learning of Graph
  Transformation Equivariant Representations via Auto-Encoding Node-wise
  Transformations},'' in \emph{{Proceedings of the IEEE Conference on Computer
  Vision and Pattern Recognition}}, 2020, pp. 7163--7172.

\bibitem{achituve2020self}
I.~Achituve, H.~Maron, and G.~Chechik, ``{Self-Supervised Learning for Domain
  Adaptation on Point Clouds},'' in \emph{Proceedings of the IEEE/CVF Winter
  Conference on Applications of Computer Vision}, 2021, pp. 123--133.

\bibitem{zhang2019unsupervised}
L.~Zhang and Z.~Zhu, ``{Unsupervised Feature Learning for Point Cloud
  Understanding by Contrasting and Clustering Using Graph Convolutional Neural
  Networks},'' in \emph{{Proceedings of the International Conference on 3D
  Vision}}.\hskip 1em plus 0.5em minus 0.4em\relax IEEE, 2019, pp. 395--404.

\bibitem{achlioptas2018learning}
P.~Achlioptas, O.~Diamanti, I.~Mitliagkas, and L.~Guibas, ``{Learning
  Representations and Generative Models for 3D Point Clouds},'' in
  \emph{{Proceedings of the International Conference on Machine
  Learning}}.\hskip 1em plus 0.5em minus 0.4em\relax PMLR, 2018, pp. 40--49.

\bibitem{qi2017pointnet++}
C.~R. Qi, L.~Yi, H.~Su, and L.~J. Guibas, ``{PointNet++: Deep Hierarchical
  Feature Learning on Point Sets in a Metric Space},'' in \emph{Proceedings of
  the Advances in Neural Information Processing Systems}, 2017, pp. 5099--5108.

\bibitem{wang2019dgcnn}
Y.~Wang, Y.~Sun, Z.~Liu, S.~E. Sarma, M.~M. Bronstein, and J.~M. Solomon,
  ``{Dynamic Graph CNN for Learning on Point Clouds},'' \emph{{ACM Transactions
  on Graphics}}, 2019.

\bibitem{nair2010rectified}
V.~Nair and G.~E. Hinton, ``{Rectified Linear Units Improve Restricted
  Boltzmann Machines},'' in \emph{{Proceedings of the International Conference
  on Machine Learning}}, 2010, pp. 807--–814.

\bibitem{Liu06-LSP}
Y.-S. Liu, J.-C. Paul, J.-H. Yong, P.-Q. Yu, H.~Zhang, J.-G. Sun, and
  K.~Ramani, ``{Automatic Least-Squares Projection of Points onto Point Clouds
  with Applications in Reverse Engineering},'' \emph{Computer-Aided Design},
  vol.~38, no.~12, pp. 1251--1263, 2006.

\bibitem{kingma2014adam}
D.~P. Kingma and J.~Ba, ``{Adam: A Method for Stochastic Optimization},'' in
  \emph{{Proceedings of the International Conference on Learning
  Representations}}, 2015, pp. 1--13.

\bibitem{EAR2013}
H.~Huang, S.~Wu, M.~Gong, D.~Cohen-Or, U.~Ascher, and H.~Zhang, ``{Edge-Aware
  Point Set Resampling},'' \emph{ACM Transactions on Graphics}, vol.~32, pp.
  9:1--9:12, 2013.

\bibitem{gal2007pose}
R.~Gal, A.~Shamir, and D.~Cohen-Or, ``{Pose-Oblivious Shape Signature},''
  \emph{IEEE Transactions on Visualization and Computer Graphics}, vol.~13,
  no.~2, pp. 261--271, 2007.

\bibitem{corsini2012efficient}
M.~Corsini, P.~Cignoni, and R.~Scopigno, ``{Efficient and Flexible Sampling
  with Blue Noise Properties of Triangular Meshes},'' \emph{IEEE Transactions
  on Visualization and Computer Graphics}, vol.~18, no.~6, pp. 914--924, 2012.

\end{thebibliography}
% \newpage
\vskip 0pt plus -1fil
\begin{IEEEbiography}[{\includegraphics[width=1in,height=1.25in,clip,keepaspectratio]{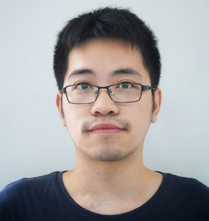}}]{Xinhai Liu}
    received the B.S. degree in computer science and technology from the Huazhong University of Science and Technology, China, in 2017. He is currently the PhD student with the School of Software, Tsinghua University. His research interests include deep learning, 3D shape analysis and pattern recognition.
\end{IEEEbiography}

\begin{IEEEbiography}[{\includegraphics[width=1in,height=1.25in,clip,keepaspectratio]{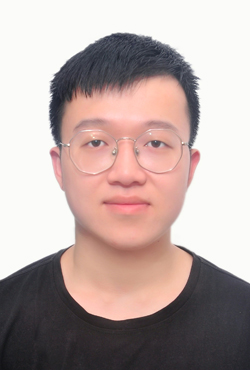}}]{Xinchen Liu}
 received the B.S. degree in Software Engineering from Hefei University of Technology, China, in 2019. He is currently the master student with school of Software, Tsinghua University. His research interests include deep learning, point cloud completion and upsampling.
\end{IEEEbiography}

\begin{IEEEbiography}[{\includegraphics[width=1in,height=1.25in,clip,keepaspectratio]{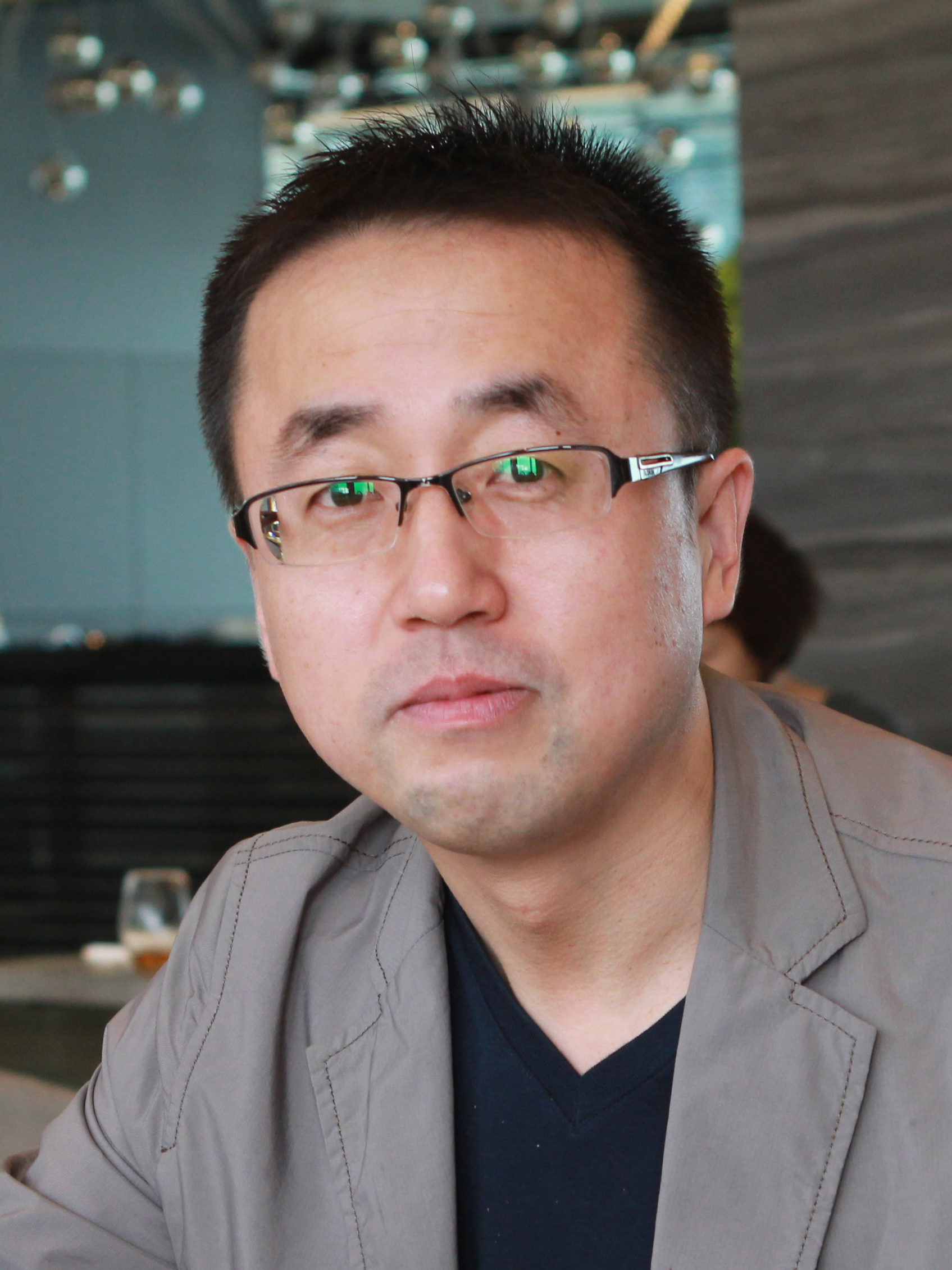}}]{Yu-Shen Liu}
    (M'18) received the B.S. degree in mathematics from Jilin University, China, in 2000, and the Ph.D. degree from the Department of Computer Science and Technology, Tsinghua University, Beijing, China, in 2006. From 2006 to 2009, he was a Post-Doctoral Researcher with Purdue University. He is currently an Associate Professor with the School of Software, Tsinghua University. His research interests include shape analysis, pattern recognition, machine learning, and semantic search.
\end{IEEEbiography}

\begin{IEEEbiography}[{\includegraphics[width=1in,height=1.25in,clip,keepaspectratio]{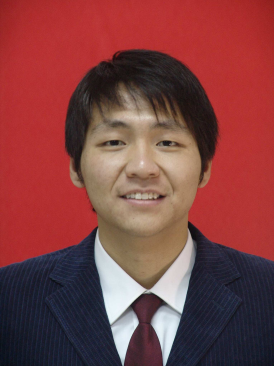}}]{Zhizhong Han}
    received the Ph.D. degree from Northwestern Polytechnical University, China, 2017. He was a Post-Doctoral Researcher with the Department of Computer Science, at the University of Maryland, College Park, USA. Currently, he is an Assistant Professor of Computer Science at Wayne State University, USA. His research interests include 3D computer vision, digital geometry processing and artificial intelligence.
\end{IEEEbiography}
\end{document}